\documentclass[letterpaper,journal]{IEEEtran}

\usepackage{amsmath,amsfonts,amssymb}
\usepackage{algorithmic}
\usepackage{algorithm}
\usepackage{array}
\usepackage{textcomp}
\usepackage{stfloats}
\usepackage{placeins}
\usepackage{url}
\usepackage{verbatim}
\usepackage{graphicx}
\usepackage{cite}
\usepackage{makecell}
\usepackage{multirow}
\usepackage{xcolor}
\usepackage{enumitem}
\usepackage{pifont}
\usepackage{booktabs}
\usepackage[caption=false,font=footnotesize]{subfig}
\usepackage{orcidlink}
\usepackage{hyperref}
\usepackage{cleveref}

\hypersetup{hidelinks}

\definecolor{my_green}{RGB}{51,102,0}
\definecolor{my_red}{RGB}{204,0,0}
\renewcommand{\checkmark}{\textcolor{my_green}{\ding{51}}}
\newcommand{\crossmark}{\textcolor{my_red}{\ding{55}}}
\providecommand{\Description}[1]{}

\begin{document}

\title{Advancing MLLM-based UAV Image Understanding and Reasoning: A Benchmark and a Training-Free Multi-Agent System}

\author{Haoyu Zhang~\orcidlink{0009-0007-9819-311X}, Shuoxun Zhang, Peng Ye, Lin Zhang, Jiakang Yuan, \\
Shenghong Yi, Yuening Wang, Tao Chen~\orcidlink{0000-0002-0779-9818},~\IEEEmembership{Senior Member,~IEEE}%
\thanks{This work was supported by the National Key R\&D Program of China (No. 2026YFE0101200) and the Shanghai Natural Science Foundation (No. 23ZR1402900). The computations in this research were performed using the CFFF platform of Fudan University. Haoyu Zhang, Shuoxun Zhang, and Peng Ye contributed equally to this work. Corresponding author: Tao Chen.}%
\thanks{\raggedright
Haoyu Zhang, Shuoxun Zhang, Lin Zhang, Jiakang Yuan and Shenghong Yi are with the Embedded Deep Learning and Visual Analysis Laboratory, College of Future Information Technology, Fudan University, Shanghai 200433, China (e-mail: 25113090081@m.fudan.edu.cn; 24210720324@m.fudan.edu.cn; 22110720068@m.fudan.edu.cn; jkyuan22@m.fudan.edu.cn; 22307130187@m.fudan.edu.cn).}%
\thanks{Tao Chen is with the Embedded Deep Learning and Visual Analysis Laboratory, College of Future Information Technology, Fudan University, Shanghai 200433, China, and also with Shanghai Innovation Institute, Shanghai 200231, China (e-mail: eetchen@fudan.edu.cn).}
\thanks{Peng Ye is with the Embedded Deep Learning and Visual Analysis Laboratory, College of Future Information Technology, Fudan University, Shanghai 200433, China, also with Shanghai Artificial Intelligence Laboratory, Shanghai 200232, China, and also with The Chinese University of Hong Kong, Hong Kong (e-mail: 20110720039@fudan.edu.cn).}%
\thanks{Yuening Wang is with the College of Future Information Technology, Fudan University, Shanghai 200433, China (e-mail: 23307130457@m.fudan.edu.cn).}%
% TODO: Add affiliations and e-mail addresses for Shuoxun Zhang and Yuening Wang.
}

\maketitle

\begin{abstract}
Multimodal Large Language Model (MLLM)-based UAV aerial image understanding and reasoning is essential for aerial intelligence yet poses distinct challenges arising from extreme scale variation, arbitrary camera orientations, and high object density. Despite growing interest, existing evaluations remain fragmented across individual datasets and narrow tasks, leaving a critical gap in unified assessment of UAV understanding and reasoning capabilities. To fill this gap, we construct UAVQA-Bench, a benchmark of 1,500 human-annotated QA pairs drawn from 13 public UAV datasets, covering 6 capability dimensions and 16 tasks in both multiple-choice and visual grounding formats. Systematic evaluation of a broad range of open-source and closed-source MLLMs as well as agent-based systems on UAVQA-Bench identifies three key failure modes: domain-toolset mismatch, unchecked error propagation, and static reasoning. Motivated by these findings, we propose UAV-MAS, a training-free multi-agent system for MLLM-based UAV aerial image understanding and reasoning, comprising a Domain-Specific Perception Engine (DSPE) that routes queries to task-appropriate visual tools, a Context-Aware Iterative Refinement module (CAIR) that validates intermediate reasoning to curb error accumulation, and a Difficulty-Aware Adaptive Search mechanism (DAAS) that adjusts search depth to question difficulty. UAV-MAS with a 32B open-source MLLM achieves 77.0\% overall accuracy on UAVQA-Bench, surpassing Gemini 3 Pro by 4.0\%, while the 8B variant improves 8.7\% over its base model.
\end{abstract}

\begin{IEEEkeywords}
UAV aerial image analysis, multi-agent systems, multimodal large language models, visual reasoning
\end{IEEEkeywords}

\section{Introduction}

% % Unmanned Aerial Vehicles (UAVs) have evolved into indispensable assets across a spectrum of critical applications, from disaster rescue and precision agriculture to urban traffic patrol and infrastructure inspection. In these high-stakes scenarios, the ability to accurately understand and reason about complex environments serves as the cornerstone of autonomous operation. Although UAVs provide a unique bird's-eye view that enables comprehensive situational awareness, this perspective introduces a formidable set of challenges: extreme scale variations resulting from altitude fluctuations, arbitrary object orientations inconsistent with ground-level priors, and a prevalence of tiny, densely packed targets amidst cluttered backgrounds. Consequently, advancing MLLM-based UAV aerial image understanding and reasoning is not merely an optimization task but a fundamental prerequisite for reliable aerial intelligence.
Unmanned Aerial Vehicles (UAVs) have evolved into indispensable assets across a spectrum of critical applications, from disaster rescue and precision agriculture to urban traffic patrol and infrastructure inspection. 
In these high-stakes scenarios, the ability to accurately perceive and interpret complex environments is essential for supporting sophisticated decision-making. 
Although UAVs provide a unique bird's-eye view that enables comprehensive situational awareness, this perspective introduces a formidable set of perceptual challenges: extreme scale variations resulting from altitude fluctuations, arbitrary object orientations inconsistent with ground-level priors, and a prevalence of tiny, densely packed targets amidst cluttered backgrounds. Consequently, advancing UAV visual perception to handle these intricate dynamics is not merely an optimization task but a fundamental prerequisite for reliable aerial artificial intelligence.

\begin{figure}[t]
  \begin{center}
    \includegraphics[width=1\columnwidth]{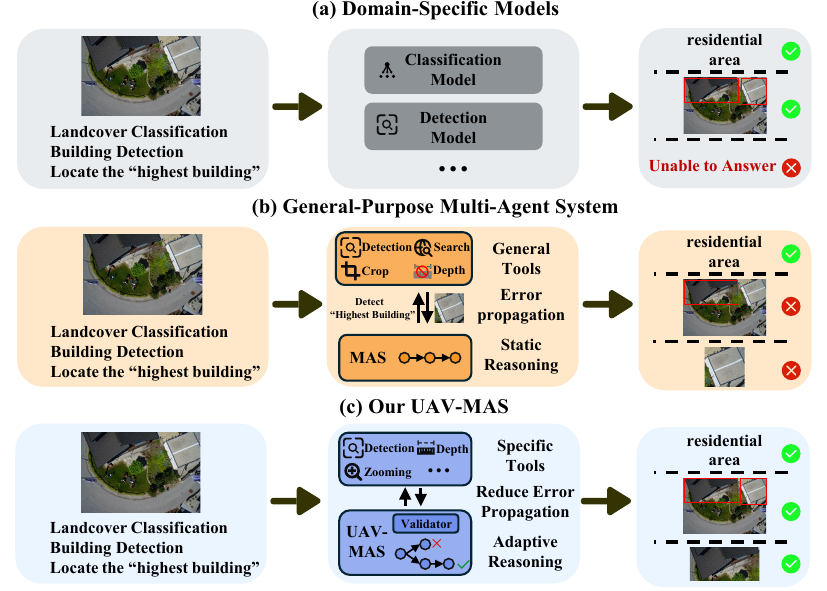}
    \caption{Comparison of MLLM-based UAV aerial image understanding and reasoning paradigms. (a) \textbf{Domain-Specific Models} handle narrow tasks but fail on complex combinatorial queries. (b) \textbf{General-Purpose MAS} suffer from toolset domain gaps and error propagation in linear static reasoning. (c) \textbf{UAV-MAS (Ours)} addresses all three failure modes via DSPE, CAIR, and DAAS.}
    \Description{Comparison of three UAV aerial image understanding and reasoning paradigms: domain-specific models, general multi-agent systems, and the proposed UAV-MAS with DSPE, CAIR, and DAAS.}
    \label{fig:fig_1}
  \end{center}
\end{figure}

Despite significant progress,
% in their training distributed domains, 
current UAV perception methods (for example, UAV-DETR~\cite{zhang2025uav} for detection and SMDE~\cite{madhuanand2021self} for depth estimation) remain confined to their respective task domains. As illustrated in Figure~\ref{fig:fig_1}(a), these task-specific models exhibit two critical limitations when facing comprehensive scenarios. First, they fail to autonomously handle complex combinatorial tasks, as they lack the mechanisms to integrate fragmented visual cues across multi-stage processes such as \textit{``Locating the highest building''}, which requires sequentially binding depth estimation with object detection. Second, they are completely devoid of the cognitive capacity for high-level reasoning and decision-making, rendering them powerless against open-ended problems like \textit{``Assessing if a specific area is suitable for emergency landing''}.
%Essentially, existing approaches perform exceptionally well as specialized sensors, yet they have not developed the ``eyes'' for holistic perception, nor do they possess the ``brain'' required for logical planning.
%As the development of Multimodal Large Language Models (MLLMs) and Multi-Agent Systems (MAS), they may be promising solutions for UAV visual perception and reasoning.
Generally speaking, existing approaches perform well as specialized sensors but lack the holistic 'eyes' and logical 'brain' for complex tasks.

%\textcolor{red}{\textbf{[TODO: @collaborator --- Write Para 2 here. Discuss the gap in existing UAV benchmarks: fragmented evaluations across narrow datasets, lack of unified assessment, and why a comprehensive benchmark is needed.]}}
\begin{table*}[ht!]
\centering
\caption{Comparison of UAVQA-Bench with existing benchmarks. 
\textbf{\#Tasks} represents the total number of subtasks.
%FAVOR-Bench covers wide video types (Third-Person, Ego-Centric, Simulation) while focusing on fine-grained motion understanding. 
%Moreover, FAVOR-Bench provides comprehensive evaluation, including close-ended QA and open-ended tasks (both GPT-assisted and a novel LLM-Free evaluation).
\textbf{Region-Level} refers to tasks with precise location annotations, e.g., bounding boxes. And \textbf{Relation} refers to relations between region-level objects.
$^*$ denotes human verification after MLLM annotation generation.
Rule-based denotes automatic generation from structured metadata without AI models.
UAVQA-Bench covers wide task scopes and provides comprehensive fully human-annotated questions, options, answers and grounding bounding boxes for the evaluation of MLLMs in the UAV domain.
}
\label{tab:comparison_otherbenchmarks}
% \vspace{1mm}
\vspace{-3mm}
\resizebox{0.9\linewidth}{!}{
\renewcommand{\arraystretch}{1.0} % Slightly reduced row height
\begin{tabular}{lccccccc}
\toprule
\raisebox{-3pt}{\multirow{2}{*}{\textbf{Benchmarks}}} & \raisebox{-3pt}{\multirow{2}{*}{\textbf{Domain}}} & \raisebox{-3pt}{\multirow{2}{*}{\textbf{Response Format}}} & \raisebox{-3pt}{\multirow{2}{*}{\textbf{\#Tasks}}} &\multicolumn{3}{c}{\textbf{Scope}} & \raisebox{-3pt}{\multirow{2}{*}{\textbf{Annotation}}}   \\
\cmidrule{5-7} 
 & & & & \textbf{Scene-Level} & \textbf{Region-Level} & \textbf{Relation} &  \\

\midrule
MMBench~\cite{liu2024mmbench} & General & QA & 20 & \checkmark & \crossmark & \checkmark & Human \\
RSVQA~\cite{lobry2020rsvqa} & Remote Sensing & QA & 5 & \checkmark & \crossmark & \crossmark & Rule-based \\
VRSBench~\cite{li2024vrsbench} & Remote Sensing & QA & 12 & \checkmark & \checkmark & \crossmark & GPT-4V$^*$ \\
XLRS-Bench~\cite{wang2025xlrs} & Remote Sensing & QA & 16 & \checkmark & \checkmark & \checkmark & GPT-4o$^*$\\
VisDrone~\cite{du2019visdrone} & UAV & Bounding Box & 4 & \crossmark & \checkmark & \crossmark & Human \\
UAVDT~\cite{du2018uavdt} & UAV & Bounding Box & 3 & \crossmark & \checkmark & \crossmark & Human \\
UrbanVideo-Bench~\cite{zhao2025urbanvideobench} & UAV & QA  & 16 & \checkmark & \crossmark & \crossmark & Gemini-1.5$^*$\\

\midrule
\textbf{UAVQA-Bench} & \textbf{UAV} & \textbf{QA \& Bounding Box} & \textbf{16} & \checkmark & \checkmark & \checkmark & \textbf{Human}  \\
\bottomrule
\end{tabular}
}
\vspace{-3mm}
\end{table*}

%However, the field still lacks a unified benchmark for evaluating UAV aerial image understanding and reasoning. Existing multimodal benchmarks are largely built on common images, where the distinctive challenges of UAV observations such as extreme scale variation, oblique viewpoints, and dense spatial layouts are largely underrepresented. Remote sensing benchmarks, while visually related, mainly target satellite-style or nadir-view interpretation and therefore do not adequately capture the perception and reasoning demands of low-altitude UAV scenarios. Meanwhile, canonical UAV datasets such as VisDrone~\cite{du2019visdrone} and UAVDT~\cite{du2018uavdt} focus primarily on narrow perception primitives such as detection and tracking, whereas recent UAV-oriented QA benchmarks~\cite{zhao2025urbanvideobench} still provide only partial coverage of the problem space, often lacking region-level grounding, inter-object relational reasoning, or rigorous fully human annotation. As summarized in Table \ref{tab:comparison_otherbenchmarks}, current evaluations remain fragmented across incompatible task settings and rely heavily on automated annotation pipelines that compromise data quality. making it difficult to assess whether a model can jointly perceive, localize, and reason in realistic aerial environments. %This fragmentation motivates the need for a comprehensive, reliable, and reproducible benchmark tailored specifically to UAV aerial intelligence.

%先一句开始
To overcome these limitations, recent Multimodal Large Language Models (MLLMs) offer a promising direction.
% The rapid evolution of Multimodal Large Language Models (MLLMs) and Multi-Agent Systems (MAS) offers a promising paradigm.
To advance MLLM-based UAV aerial image understanding and reasoning, we first introduce UAVQA-Bench, a comprehensive and fully human-annotated benchmark. As summarized in Table \ref{tab:comparison_otherbenchmarks}, previous comprehensive multimodal benchmarks are largely built on general and remote sensing images, while existing UAV benchmarks still provide only partial coverage of the problem space. In addition, they often rely heavily on automated annotation pipelines, which may compromise data quality.
As a comparison, our UAVQA-Bench covers 6 capability dimensions and 16 tasks over 1,500 carefully curated samples drawn from 13 diverse public UAV datasets, and adopts a closed-ended, objectively scorable evaluation protocol that enables reliable and reproducible assessment. All questions, candidate options, answers, and grounding annotations are produced and verified by human annotators, ensuring higher correctness, consistency, and contextual faithfulness. 

With the dataset established, a systematic evaluation of a broad range of MLLMs and MLLM-based multi-agent systems is conducted. Initial findings suggest that these methods are generally capable of processing diverse problem types within UAV scenarios, demonstrating an acceptable baseline in general aerial image understanding and reasoning. 
% More specifically, 
In detail, closed-source methods show an overall performance advantage. Among open-source methods, larger models generally perform better in aggregate, whereas smaller models can still be stronger on specific tasks such as visual grounding.
Despite this functional baseline, our further analysis
% With the dataset established, systematic evaluation of a broad range of MLLMs and MLLM-based multi agentsystems are conducted and 
reveals three recurring failure modes as shown in Figure \ref{fig:fig_1}(b): (\textit{i}) \textbf{domain-toolset mismatch}, where standard vision tools trained on ground-level data degrade on aerial patterns; (\textit{ii}) \textbf{unchecked error propagation}, where early-stage tool errors cascade through multi-step chains without self-correction; and (\textit{iii}) \textbf{static reasoning}, where fixed linear strategies fail to adapt to varying aerial task complexity. 

To address the above challenges, we propose UAV-MAS, a training-free multi-agent system for MLLM-based UAV aerial image understanding and reasoning. UAV-MAS integrates three targeted components. First, the Domain-Specific Perception Engine (DSPE) equips the system with specialized aerial operators and a precise tool selection mechanism to overcome domain-toolset mismatch on UAV imagery. Second, the Context-Aware Iterative Refinement (CAIR) strategy introduces hierarchical verification that actively detects and corrects reasoning errors, preventing noise accumulation in multi-step inference. Third, the Difficulty-Aware Adaptive Search (DAAS) mechanism dynamically adjusts reasoning level based on task complexity, balancing computational efficiency with thoroughness. Experiments demonstrate strong robustness and reasoning accuracy: UAV-MAS with 32B MLLMs achieves 4.0\% higher Overall Accuracy on UAVQA-Bench than the closed-source Gemini 3 Pro, while relying solely on training-free open-source models.\par

In summary, our main contributions are:
%\begin{itemize}[itemsep=1pt, topsep=2pt, parsep=0pt, partopsep=0pt]
\begin{itemize}
  \item We introduce UAVQA-Bench, a comprehensive, fully human-annotated benchmark for UAV aerial image understanding and reasoning. It covers 6 capability dimensions across 16 tasks (in multiple-choice and visual grounding formats), with 1,500 samples drawn from 13 diverse public UAV datasets. We systematically evaluate a broad range of open/closed-source MLLM-based methods on UAVQA-Bench.
  \item We propose UAV-MAS, a training-free multi-agent system for MLLM-based UAV aerial image understanding and reasoning. It integrates a Domain-Specific Perception Engine (DSPE) for aerial-adapted tool use, a Context-Aware Iterative Refinement (CAIR) strategy for error-resilient multi-step reasoning, and a Difficulty-Aware Adaptive Search (DAAS) mechanism for complexity-adaptive inference.
  \item Extensive experiments show that UAV-MAS achieves state-of-the-art performance, surpassing the powerful closed-source Gemini 3 Pro by 4.0\% in Overall Accuracy on UAVQA-Bench using only training-free 32B MLLMs, while offering a more efficient accuracy--cost trade-off than brute-force test-time scaling at comparable accuracy.
\end{itemize}

\section{Related Work}
\subsection{UAV Aerial Image Analysis}
Recent advances in deep learning have spurred the development of specialized models tailored for diverse analysis tasks in the UAV domain, achieving impressive results.
%The small-scale perception model in the drone field is a lightweight artificial intelligence model optimized specifically for drone embedded platforms. It can process visual and other sensor data in real-time with limited computing power, performing tasks 
These tasks include detection\cite{chakrabarty2026yolo26,zhang2025uav,gao2025dfir,khanpour2025uav,chen2025globallocal,xi2024dedet}, counting\cite{maimaitijiang2026wheatai,kharismawati2025maizestandcounting,youme2025panoptic}, depth estimation\cite{madhuanand2021self,chen2025codrone,lin2025generalization,shi2024precise}. Representative detection methods include UAV-DETR\cite{zhang2025uav}, an end-to-end detector tailored to small objects in drone images; global--local feature fusion with semantic guidance\cite{chen2025globallocal}; and detection-oriented exposure correction for nighttime drone views\cite{xi2024dedet}. Recent methods such as FLDet~\cite{wang2025fldet} and TGCADNet~\cite{sun2026tgcadnet} respectively emphasize lightweight detection and text-guided contextual modeling for small objects in UAV scenes. Although effective under their respective settings, these methods remain optimized for predefined categories and specific perception tasks.
%In depth estimation, CoDrone\cite{chen2025codrone} adopts an edge cloud collaborative framework that combines depth estimation and visual language models to enhance drone navigation capabilities, and optimizes task scheduling through a reinforcement learning scheduler. However, the system is designed specifically for a single navigation task and cannot process other perception tasks.\par
Rex-Omni~\cite{jiang2025rexomni} leverages the strong generalization of MLLMs to enable open-world object detection. It reformulates detection as next-point prediction with quantized coordinates and achieves competitive zero-shot performance. For depth estimation, Shi et al.~\cite{shi2024precise} recover scene depth from consecutive UAV frames by combining affine patch transformations with geometric constraints, targeting the narrow-baseline conditions common in aerial sequences.

%Although the aforementioned work performs well, there are still two problems: firstly, a single model can only handle a single task and cannot complete multiple tasks.  Second, the model cannot solve tasks that need understanding and thinking, such as VQA.\par

%Despite these advances, existing UAV perception models remain largely confined to narrow, isolated tasks and are unable to perform the kind of integrated, multi-step reasoning required in real-world autonomous operations. They excel as specialized components but fall short in combining diverse perceptual capabilities to support complex, open-ended instructions or adaptive decision-making in unstructured environments.
Despite these advances, existing UAV aerial image analysis models remain confined to isolated tasks and cannot integrate diverse visual capabilities for multi-step reasoning or adaptive decision-making in real-world scenarios.

\subsection{UAV Aerial Image Understanding Benchmarks}

% \textcolor{red}{\textbf{TODO:} Survey the current data landscape for UAV aerial image understanding and reasoning benchmarks. Cover existing datasets (scale, annotation types, task diversity, limitations) and justify why a new comprehensive benchmark is needed.}

Existing benchmarks for evaluating UAV aerial image understanding and reasoning exhibit notable limitations. Typical datasets such as VisDrone \cite{du2019visdrone, cao2021visdrone} and DOTA \cite{xia2018dota} primarily serve narrow perception primitives like detection and tracking, falling short of assessing higher-order cognitive capabilities. UAVScenes \cite{wang2025uavscenes} extends to multi-modal perception and six task types including segmentation, yet its focus remains on perceptual primitives rather than comprehensive reasoning. RSVQA~\cite{lobry2020rsvqa} provides rule-generated QA for satellite and aerial orthophotos, but does not target low-altitude UAV scenes. More recent remote-sensing benchmarks, such as VRS-Bench \cite{li2024vrsbench} and XLRS-Bench \cite{wang2025xlrs}, remain dominated by nadir-view perspectives and task designs that fail to capture extreme scale variation, oblique viewpoints, and dense spatial layouts.

Recent benchmarks including UrbanVideo-Bench \cite{zhao2025urbanvideobench} have attempted to construct more comprehensive evaluation frameworks. However, they generally suffer from incomplete task coverage and incompatible settings, along with reliance on automated annotation pipelines, resulting in data quality that falls short of human-annotated standards. %These limitations collectively indicate that existing benchmarks are constrained in terms of task dimensionality, perspective adaptation, data quality, and evaluation capability, making it difficult to effectively assess an agent’s ability to jointly perceive, localize, and reason in authentic aerial environments.

\subsection{Multi-Agent System}
%Multi-agent systems, a research field focusing on solving complex tasks through interaction and collaboration among multiple autonomous agents, has received widespread attention in academia and industry in recent years, yielding a plethora of research outcomes with theoretical and practical value
Multi-agent systems powered by MLLMs have recently gained traction as a framework for tackling complex, multi-step visual reasoning tasks, where collaborative agents decompose high-level queries and coordinate visual analysis with reasoning~\cite{li2025dyfo,zhao2025pyvision,scofield2026multi,ke2026mas,rodriguez2026agent,wang2025internvl3,ma2026paper2rebuttal,adimulam2026orchestration,lee2026motion,hui2026agentgc}. ReAct~\cite{yao2022react} introduces a prompting framework that interleaves reasoning and actions, enabling language models to reason and act jointly. It can reduce hallucination and improve performance in tasks like question answering and decision-making. DyFo~\cite{li2025dyfo} adopts a static toolkit architecture and builds a multimodal interaction framework based on ReAct, and utilizes the MCTS search algorithm to achieve a method that enhances the fine-grained visual understanding capability of large-scale multimodal models without training. %However, the linear reasoning paradigm of ReAct lacks self-correcting capabilities, and DyFo's singular tool cannot accomplish tasks that require specialized external tools.
PyVision\cite{zhao2025pyvision} adopts the Model Context Protocol (MCP) toolkit as the interaction interface, follows the ReAct paradigm to implement iterative cycles of multi-step reasoning and tool invocation, and utilizes the MCTS search algorithm to enhance the adaptability and interpretability of complex visual reasoning tasks. %However, its dynamic MCP toolkit results in excessively high model size requirements. \par

%Although these methods perform well in their respective fields, there are still significant limitations when generalizing to drone related tasks. Including issues such as inability to correct inference noise, tool incompatibility, and high model size requirements. \par

%Although these methods perform well in their respective fields, they are ill-suited for UAV perception due to three key limitations: reliance on standard vision tools that falter under aerial imaging conditions, error propagation from tool chaining without self-correction, and inflexible planning strategies that cannot adaptively balance reasoning depth with task complexity. These shortcomings hinder robust, end-to-end autonomy in real-world drone operations—motivating the need for a perception-aware, resilient, and dynamically adaptive multi-agent framework tailored to the aerial domain.
Although effective in their domains, these methods are ill-suited for UAV aerial image understanding due to ground-biased vision tools, error propagation from tool chaining, and rigid planning that fails to adapt to aerial task complexity---motivating the need for a resilient and dynamically adaptive multi-agent framework tailored to the aerial domain.

\section{UAVQA-Bench}

\begin{figure*}[!t]
\centering
\includegraphics[width=1\linewidth]{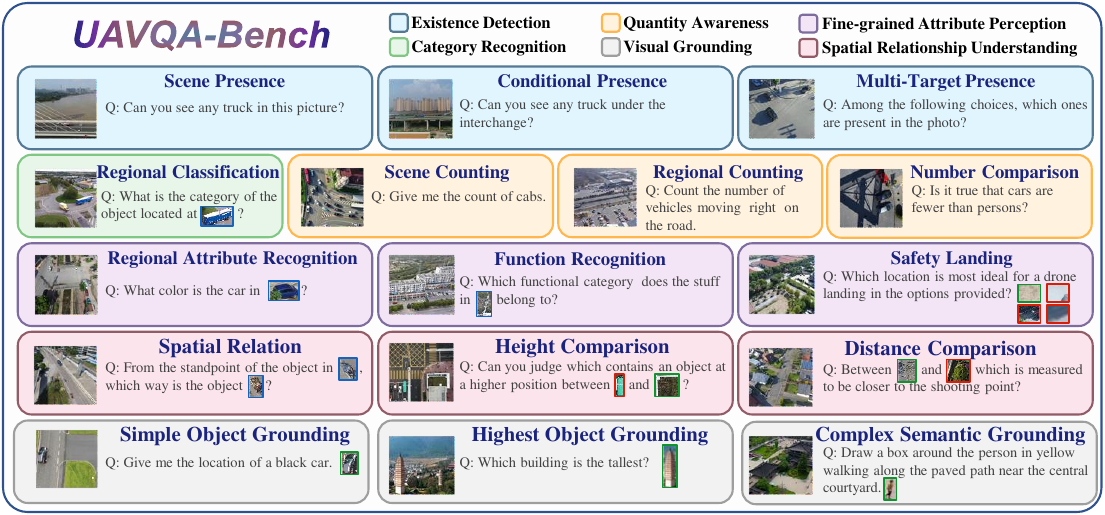}
\caption{Overview of UAVQA-Bench. The benchmark assesses 6 key capabilities through 16 distinct tasks, collectively forming a comprehensive evaluation framework.}
% \Description{Overview of UAVQA-Bench capabilities and 16 tasks used for evaluating UAV visual reasoning.}
\label{benchmark}
\end{figure*}

To address the limitations in existing UAV aerial image understanding and reasoning evaluation platforms, we introduce ``UAVQA-Bench'', a comprehensive benchmark covering multi-task, multi-scenario, and multi-scale settings.
As shown in~\cref{tab:comparison_otherbenchmarks}, UAVQA-Bench features comprehensive scopes including scene-level, region-level and relation between regional objects, which ensures a holistic evaluation beyond the scope of existing UAV datasets. All annotations are provided by highly educated human annotators under an inspection-revision process. Compared to synthetic annotations in benchmarks like VRS-Bench~\cite{li2024vrsbench} and UrbanVideo-Bench~\cite{zhao2025urbanvideobench}, human-crafted annotations demonstrate superior quality in terms of correctness, consistency, and contextual understanding over synthetic outputs from large models.
UAVQA-Bench assesses 6 key capabilities of UAV agent systems and encompasses 16 distinct tasks, systematically examining UAV agents' multi-dimensional understanding and reasoning abilities in complex environments.
Notably, our evaluation adopts a closed-ended QA (multiple-choice) and grounding box matching approach, which simplifies assessment and enhances reproducibility by eliminating the ambiguity inherent in open-ended responses, while ensuring reliable evaluation through standardized answer spaces.

\subsection{Dataset Construction and Data Sources}
UAVQA-Bench comprises 1,500 samples, each pairing a high-quality UAV image with a carefully constructed question--answer instance. To ensure task relevance and visual diversity, we manually selected images from 13 publicly available UAV datasets: AU-AIR~\cite{bozcan2020air}, WebUAV-3M~\cite{zhang2022webuav}, VisDrone-DET2019~\cite{du2019visdronedet2019}, Semantic Drone~\cite{semantic_drone_dataset}, DroneVehicle~\cite{sun2022dronevehicle}, UAVDT~\cite{du2018uavdt}, VDD~\cite{cai2025vdd}, UDD~\cite{chen2018udd}, UAVid~\cite{lyu2020uavid}, WildUAV~\cite{florea2021wilduav}, HazyDet~\cite{feng2024hazydet}, AnimalDrone~\cite{zhu2021animaldrone}, and UAV123~\cite{2016uav123}, with their task-level provenance summarized in~\cref{tab:datasource_app}. Dataset construction followed a multi-stage quality-control process involving seven volunteers with backgrounds in computer vision: two selected suitable images and annotated bounding boxes where necessary, two constructed the questions, answer options, and ground-truth answers, and the remaining three independently reviewed the image quality, bounding-box annotations, question and option quality, and answer correctness. For tasks requiring distractor options, three distractors were selected from six LLM-generated candidates. Only samples receiving unanimous approval were retained; the remaining samples were revised or removed, followed by random spot checks of the resulting dataset.

\begin{table*}[!th]
\centering
\caption{Data sources of each task.}
\label{tab:datasource_app}
\small % 适当增大字体，提高表格可读性
\setlength{\tabcolsep}{5pt} % 调整列间距
\renewcommand{\arraystretch}{1.2} % 增加行高，避免文本堆叠

% 使用 @{}p{...} 定义列宽，并使用 | 增加 L3 和 Template 之间的垂直线
\resizebox{0.99\linewidth}{!}{
\begin{tabular}{l|l|l}
%{@{}p{2.5cm}|p{9.5cm}|p{3cm}@{}}
\toprule
\textbf{Capability} & \textbf{Task} & \textbf{Data Source} \\
\hline

% ----- Perception -----
% \multicolumn{3}{@{}l}{\textbf{Perception Tasks}} \\ % L1/L2 分组标题
% \midrule

\multirow{3}{*}{Existence Detection} & Scene Presence & WebUAV-3M, VisDrone-DET2019, Semantic Drone \\
\cline{2-3} 
& Conditional Presence & AU-AIR, WebUAV-3M, Semantic Drone\\
\cline{2-3} 
& Multi-Target Presence & AU-AIR, WebUAV-3M, VisDrone-DET2019, Semantic Drone, UAVDT, HazyDet \\

\hline

\multirow{1}{*}{Category Recognition} & Regional Classification & AU-AIR, VisDrone-DET2019\\

\hline

\multirow{3}{*}{Quantity Awareness} & Scene Counting & AU-AIR, WebUAV-3M, Semantic Drone, AnimalDrone\\
\cline{2-3} 
& Regional Counting & AU-AIR, WebUAV-3M, VisDrone-DET2019, Semantic Drone, AnimalDrone\\
\cline{2-3} 
& Number Comparison & AU-AIR, WebUAV-3M, VisDrone-DET2019, Semantic Drone, UAVDT, HazyDet, AnimalDrone, UAV123\\

\hline

\multirow{3}{*}{\shortstack[l]{Fine-grained Attribute\\Perception}} & Regional Attribute Recognition & AU-AIR, VisDrone-DET2019, DroneVehicle\\
\cline{2-3} 
& Function Recognition & VisDrone-DET2019, VDD, UDD\\
\cline{2-3} 
& Safety Landing & VisDrone-DET2019 \\

\hline

\multirow{3}{*}{\shortstack[l]{Spatial Relationship\\Understanding}} & Spatial Relation & AU-AIR, VisDrone-DET2019, Semantic Drone \\
\cline{2-3} 
& Height Comparison & VisDrone-DET2019, UAVDT, VDD, UDD, UAV123\\
\cline{2-3} 
& Distance Comparison & AU-AIR, VisDrone-DET2019, UAVDT, VDD, UDD, UAVid, WildUAV\\

\hline

\multirow{3}{*}{Visual Grounding} & Simple Object Grounding & AU-AIR, VisDrone-DET2019\\ 
\cline{2-3} 
& Complex Semantic Grounding & AU-AIR, WebUAV-3M  \\ 
\cline{2-3} 
& Highest Object Grounding & WebUAV-3M, VDD, Semantic Drone \\ 

\bottomrule
\end{tabular}
}
\vspace{-5pt}
\end{table*}

For region-level tasks, we manually annotate the selected images with precise bounding boxes. We then design task-specific question templates with diverse linguistic expressions to reduce the risk of models exploiting superficial patterns. Candidate answers for multiple-choice questions include plausible visual or semantic distractors, while all questions, options, ground-truth answers, and grounding annotations are manually authored and rigorously cross-verified. For tasks requiring spatial outputs, such as visual grounding, coordinates are represented as $\{\mathopen{<}x_1\mathclose{>}\mathopen{<}y_1\mathclose{>}\mathopen{<}x_2\mathclose{>}\mathopen{<}y_2\mathclose{>}\}$, where $(x_1,y_1)$ and $(x_2,y_2)$ denote the top-left and bottom-right corners, respectively. All coordinates are normalized by image width and height and rounded to three decimal places for consistent spatial representation.

% % \begin{figure}[!t]

% % \centering
% % \includegraphics[width=\linewidth]{figs/dataset3.pdf}
% % \caption{Overview of UAVQA-Bench. The benchmark assesses $6$ key capabilities through $16$ distinct tasks, collectively forming a comprehensive evaluation framework.}
% \Description{Overview of UAVQA-Bench capabilities and 16 tasks used for evaluating UAV visual reasoning.}
% \label{benchmark}
% \end{figure}

% \begin{figure*}[!t]
% \centering
% \includegraphics[width=0.85\linewidth]{figs/benchmark/dataset1-1.pdf}
% \caption{Overview of UAVQA-Bench. The benchmark assesses $6$ key capabilities through $16$ distinct tasks, collectively forming a comprehensive evaluation framework.}
% % \Description{Overview of UAVQA-Bench capabilities and 16 tasks used for evaluating UAV visual reasoning.}
% % \label{benchmark}
% % \end{figure*}

\subsection{Capability Dimensions}
As illustrated in Figure~\ref{benchmark}, UAVQA-Bench establishes 6 core capabilities comprising 16 diverse tasks, ranging from basic perception to high-level reasoning.

%\begin{itemize}[itemsep=3pt, parsep=0pt, topsep=0pt, partopsep=0pt, labelsep=5pt,leftmargin=*]
\begin{itemize}[labelsep=5pt,leftmargin=*]
    \item \textbf{Existence Detection (ED)} evaluates the agent's ability to identify object presence under varying conditions, including \textit{Scene Presence}, \textit{Conditional Presence}, and \textit{Multi-Target Presence}.

    \item \textbf{Category Recognition (CR)} focuses on fundamental semantics through \textit{Regional Classification} of detected entities.

    \item \textbf{Quantity Awareness (QA)} needs precise counting and numerical logic via \textit{Scene/Regional Counting} and \textit{Number Comparison}.

    \item \textbf{Fine-grained Attribute Perception (FAP)} assesses the agent's granular understanding of entity properties and scene affordances. It encompasses \textit{Attribute Recognition} (e.g., identifying colors and orientations), \textit{Function Recognition} of specific objects, and \textit{Safety Landing} analysis, which requires the model to evaluate surface conditions to identify secure touchdown zones for the UAV.

    \item \textbf{Spatial Relationship Understanding (SRU)} examines 3D spatial reasoning from a distinctive aerial perspective. Beyond identifying \textit{Spatial Relations} between discrete entities or their relative positioning to the UAV, this dimension includes \textit{Height Comparison} and \textit{Distance Comparison}. Here, ``higher'' denotes greater physical elevation in 3D space rather than a smaller image-plane vertical coordinate, while distance denotes the camera-to-object range. Agents must therefore reason about relative elevation and depth in complex, multi-level environments.

    \item \textbf{Visual Grounding (VG)} tests the precise alignment between complex linguistic queries and visual regions. Spanning from \textit{Simple Object Grounding} to \textit{Complex Semantic Grounding} with multi-step reasoning, this capability evaluates the agent's cross-modal comprehension. Notably, it includes \textit{Highest Object Grounding}, which challenges the model to identify and localize the vertically most prominent entities, such as the tallest building in a dense urban scene.
\end{itemize}

\begin{figure*}[!t]
  \centering
  \subfloat[Capability composition]{%
    \includegraphics[
      width=0.2185\textwidth,
      trim=0.2cm 0.5cm 0.2cm 0.2cm,
      clip
    ]{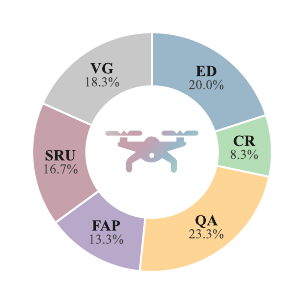}}
  \hfill
  \subfloat[Target spatial distribution]{%
    \includegraphics[width=0.2755\textwidth]{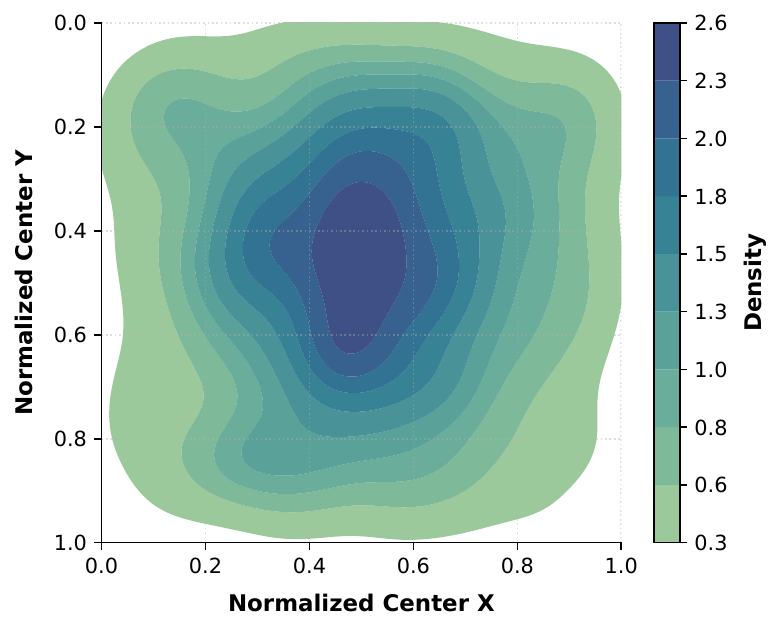}}
  \hfill
  \subfloat[High-frequency vocabulary]{%
    \includegraphics[width=0.38\textwidth]{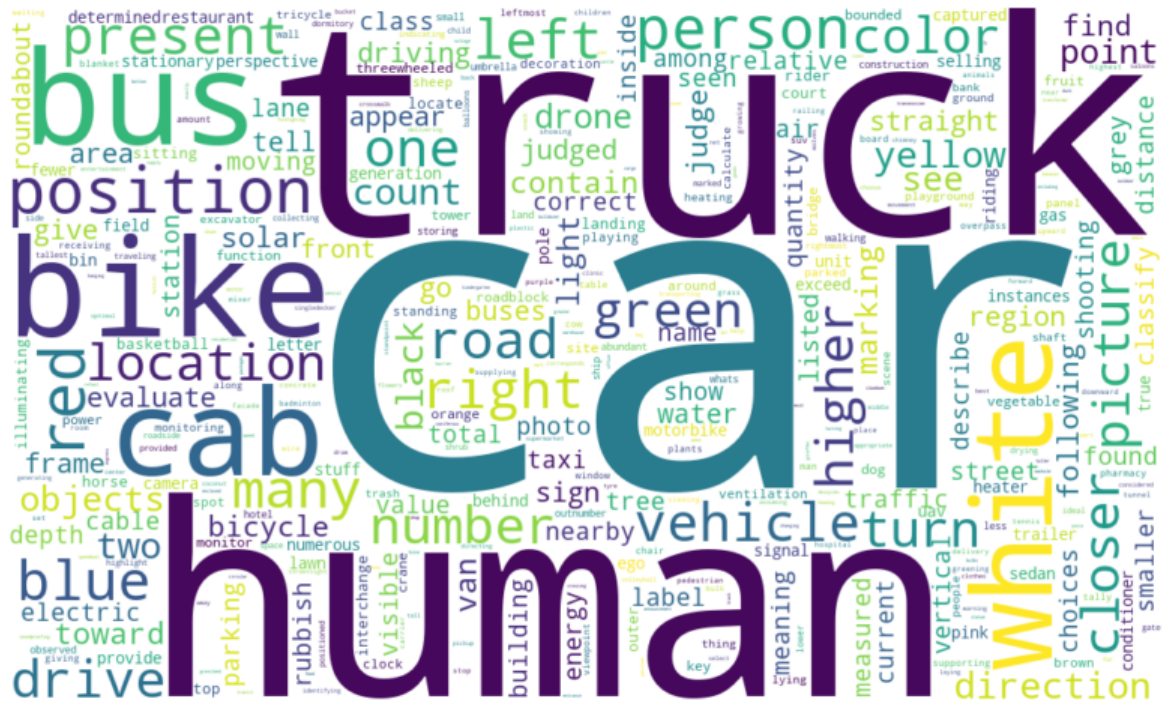}}

  \vspace{0.7em}

  \subfloat[Question token lengths]{%
    \includegraphics[width=0.2256\textwidth]{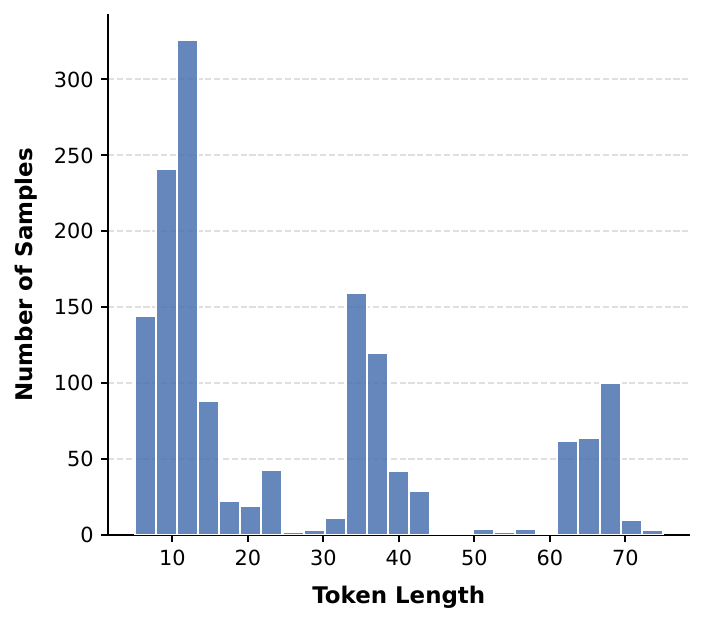}}
  \hfill
  \subfloat[Image resolutions]{%
    \includegraphics[width=0.2256\textwidth]{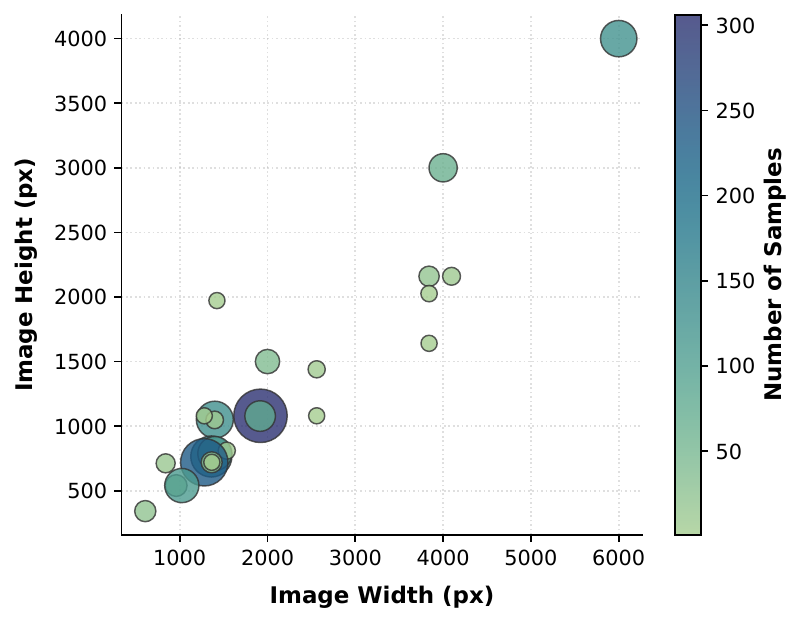}}
  \hfill
  \subfloat[Bounding-box areas]{%
    \includegraphics[width=0.2256\textwidth]{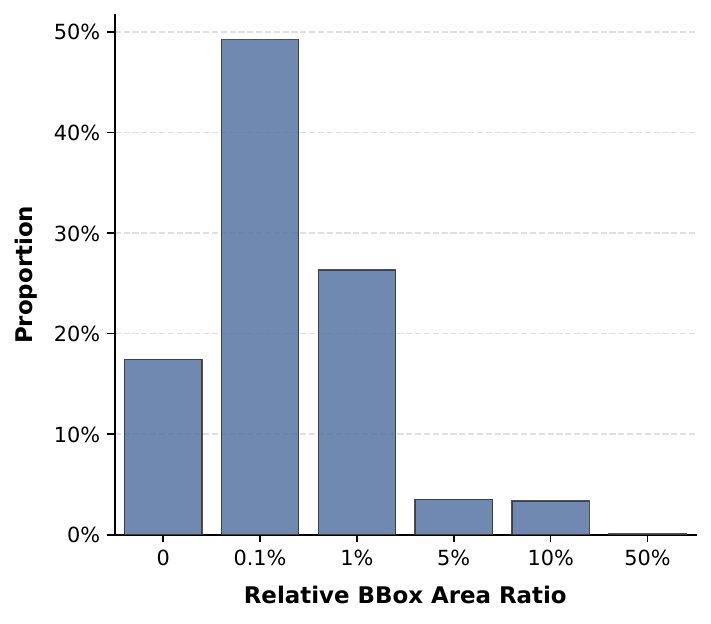}}
  \hfill
  \subfloat[Bounding-box shapes]{%
    \includegraphics[width=0.2256\textwidth]{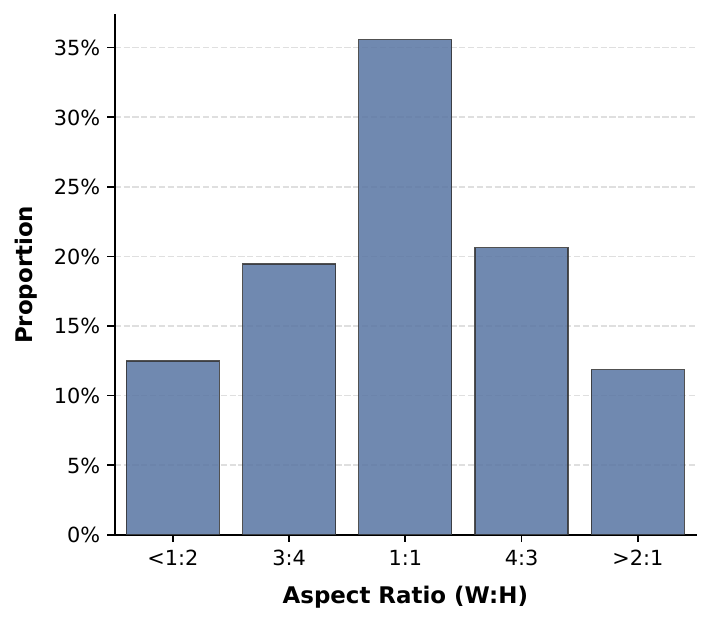}}
  \caption{Dataset statistics of UAVQA-Bench. (a) Composition of the six capability dimensions. (b) Spatial distribution of target centers. (c) High-frequency vocabulary. (d) Question token-length distribution. (e) Image-resolution distribution. (f) Bounding-box area relative to image area. (g) Bounding-box aspect-ratio distribution.}
  \label{fig:Dataset statistics}
\end{figure*}

\subsection{Evaluation Protocol and Metrics}
Most UAVQA-Bench tasks adopt a multiple-choice format, whereas visual grounding tasks require a predicted bounding box. A multiple-choice instance is counted as correct when the model selects the ground-truth option. For visual grounding, the model is explicitly prompted to return coordinates in the same normalized format used by the annotations, and a prediction is counted as correct when its Intersection over Union (IoU) with the ground-truth box reaches or exceeds $0.5$.

We report two complementary aggregate metrics: \textbf{Overall Accuracy (OA)} and \textbf{Average Accuracy (AA)}. Let $T=16$ denote the number of tasks, and let $m_i$ and $c_i$ denote the total and correctly answered samples in task $i$, respectively. The metrics are defined as
\begin{equation}
    \text{OA} = \frac{\sum_{i=1}^{T} c_i}{\sum_{i=1}^{T} m_i} \times 100\%,
\end{equation}
\begin{equation}
    \text{AA} = \frac{1}{T} \sum_{i=1}^{T} \frac{c_i}{m_i} \times 100\%.
\end{equation}
OA measures sample-weighted performance over the complete benchmark, while AA assigns equal weight to every task. Reporting both metrics therefore captures overall robustness without allowing tasks with more samples to obscure performance differences across capability dimensions.

\subsection{Dataset Statistics}

Figure~\ref{fig:Dataset statistics} provides a unified statistical overview of UAVQA-Bench. The benchmark maintains a relatively balanced composition across the six capability dimensions (Figure~\ref{fig:Dataset statistics}(a)), covering both foundational perception and more challenging reasoning-oriented tasks. Target objects are broadly distributed across the image plane rather than concentrated in a small set of locations (Figure~\ref{fig:Dataset statistics}(b)). The word cloud in Figure~\ref{fig:Dataset statistics}(c) further summarizes the benchmark's lexical coverage across object categories, attributes, spatial relationships, and actions. Non-content words, such as prepositions and formatting tokens, are removed so that font size directly reflects the relative frequency of semantically meaningful terms.

UAVQA-Bench also presents substantial diversity in both textual and visual inputs: question lengths range from short queries to long compositional reasoning prompts, while image resolutions span from low-resolution to ultra-high-resolution aerial imagery (Figure~\ref{fig:Dataset statistics}(d) and~(e)). The bounding-box statistics in Figure~\ref{fig:Dataset statistics}(f) and~(g) further reveal the fine-grained perception challenge. Specifically, 49.2\% of the boxes occupy only 0.1\%--1\% of the image area, and another 17.5\% occupy less than 0.1\%; by contrast, just 0.1\% cover more than 50\% of an image. Near-square boxes, with width-to-height ratios between 3:4 and 4:3, form the largest group, although the complete distribution spans a wide range of object shapes. Together, these statistics demonstrate diversity in capability coverage, target location and geometry, language, and image resolution, making UAVQA-Bench a challenging testbed for multimodal agents in UAV scenarios.

\FloatBarrier

\begin{figure*}[!t]
  \begin{center}
    \includegraphics[width=1\textwidth]{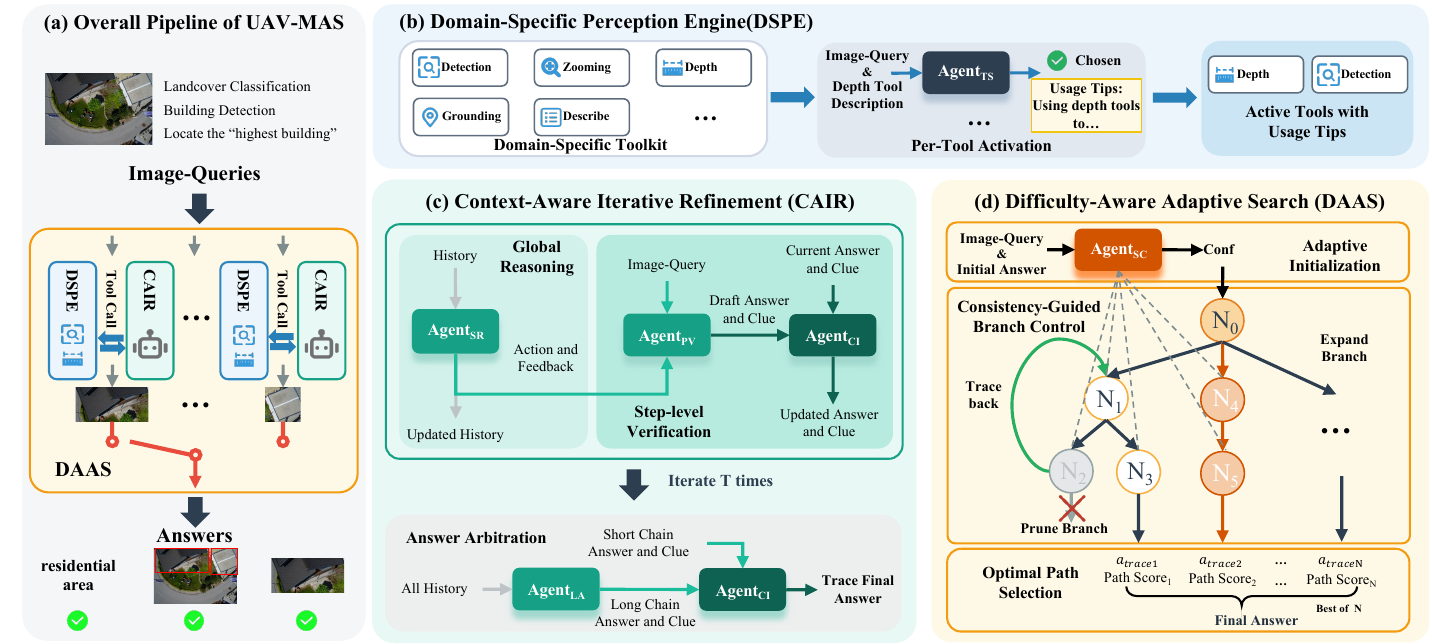}
    \caption{Overview of UAV-MAS. (a) The overall pipeline of UAV-MAS. (b) The Domain-Specific Perception Engine (DSPE) tailored for UAV perception. (c) The Context-Aware Iterative Refinement (CAIR) strategy to mitigate tool invocation noise. (d) The Difficulty-Aware Adaptive Search (DAAS) mechanism for dynamic reasoning exploration.}
    \Description{System overview of UAV-MAS including pipeline, DSPE toolkit, CAIR iterative refinement, and DAAS adaptive search.}
    \label{fig:fig_2}
  \end{center}
\end{figure*}

\section{Method}

% Systematic evaluation of a broad range of MLLMs and agent-based systems on UAVQA-Bench as shown in Table~\ref{tab:main_results} reveals three recurring failure modes: domain-toolset mismatch, unchecked error propagation, and static reasoning. To address these, we propose \textbf{UAV-MAS}, a training-free multi-agent system. As illustrated in Figure~\ref{fig:fig_2}, UAV-MAS comprises three targeted components. First, the \textbf{Domain-Specific Perception Engine (DSPE)} equips the agent with aerial-targeted perception tools, directly countering the domain-toolset mismatch between ground-level pretraining data and overhead imagery. Second, the \textbf{Context-Aware Iterative Refinement (CAIR)} strategy introduces step-level verification that detects and corrects unreliable tool feedback before it corrupts the accumulated reasoning state, preventing error propagation. Third, the \textbf{Difficulty-Aware Adaptive Search (DAAS)} mechanism derives a per-query pruning threshold from estimated query difficulty and selects the optimal reasoning path by global coherence, replacing static linear reasoning with complexity-adaptive exploration. Together, these modules enable \textbf{UAV-MAS} to achieve robust performance on aerial visual question answering without any task-specific training.
Evaluating MLLM-based methods on UAVQA-Bench (Table~\ref{tab:main_results}) exposes three key failures: domain-toolset mismatch, error propagation, and static reasoning. To address these, we propose \textbf{UAV-MAS} (Figure~\ref{fig:fig_2}), a training-free multi-agent system comprising three core modules. First, the \textbf{Domain-Specific Perception Engine (DSPE)} provides aerial-targeted tools to overcome domain mismatch. Second, \textbf{Context-Aware Iterative Refinement (CAIR)} utilizes step-level verification to correct unreliable feedback and halt error propagation. Third, \textbf{Difficulty-Aware Adaptive Search (DAAS)} replaces static reasoning with complexity-adaptive exploration by pruning paths based on query difficulty. Together, these modules enable \textbf{UAV-MAS} to achieve robust aerial image understanding and reasoning performance without task-specific training.

\subsection{Domain-Specific Perception Engine}
While general-purpose Multimodal Large Language Models (MLLMs) generalize well to common scenes, aerial images in UAV understanding and reasoning tasks expose structural challenges that standard models cannot adequately handle: extreme scale variation, arbitrary object orientation, and high-density small object clutter. Generic models also lack priors for aerial-specific reasoning subtasks (e.g., distinguishing ground-level from rooftop targets, counting tiny objects under dense occlusion). To address these gaps, we design the \textbf{Domain-Specific Perception Engine (DSPE)}: a modular suite of aerial-targeted tools combined with a per-tool activation mechanism for lightweight scheduling.

\textbf{Domain-Specific Toolkit.} The toolkit comprises 5 core tools targeting the principal visual bottlenecks in overhead imagery: \textbf{1) Context-Aware Zooming} employs a soft-boundary mechanism that retains edge context during cropping to enhance local resolution while preserving semantic continuity, particularly important for aerial targets that often span only a few pixels at typical UAV altitudes. \textbf{2) Fine-Grained Explicit Description} converts implicit visual features (e.g., textures, behaviors) into structured text via instruction following, supporting downstream reasoning. \textbf{3) Distance Estimation} utilizes Depth Anything 3~\cite{depthanything3} to generate pseudo-depth maps for reconstructing 3D structures from monocular imagery, distinguishing ground-level from rooftop targets based on relative depth discontinuities. \textbf{4) Semantic Grounding} locates individual targets based on explicit instructions (e.g., ``Locate the red vehicle'') for precise spatial positioning. \textbf{5) Open-Vocabulary Detection with De-hallucination} addresses hallucination patterns specific to dense aerial scenes through a two-stage verification. First, prior to localizing, an MLLM-based existence verification confirms whether the target category is present at all, preventing the model from forcibly generating bounding boxes for absent targets. Second, to address the issue where MLLMs systematically hallucinate bounding boxes at regular intervals during autoregressive decoding, we apply a heuristic filter. It removes a detection $c_i$ only when the equidistant condition $|c_{i+1} - 2c_i + c_{i-1}| < \delta$ holds \emph{continuously} across multiple consecutive detections, preserving legitimate near-uniform layouts (e.g., a row of parked vehicles) while suppressing fabricated repetitive patterns.

\textbf{Per-Tool Activation.} In aerial image analysis, tool selection is query-type-dependent: counting tasks require detection while spatial reasoning tasks require depth estimation rather than description. Small open-source MLLMs frequently fail when asked to jointly select and configure multiple tools from a single prompt, due to context overflow or hallucinated activations. To address this, each tool is assigned a dedicated lightweight agent ($\text{Agent}_{\text{TS}}$) responsible for a single binary decision: ``Should this tool be activated for the current image-query pair?'' If activated, $\text{Agent}_{\text{TS}}$ additionally generates the required execution parameters. 
% (e.g., crop region or detection prompt keywords).

Together, the 5 aerial-targeted tools and per-tool activation ensure that the base MLLM receives enriched, task-relevant perceptual inputs without needing to manage complex multi-tool scheduling---directly addressing the domain gap in UAV aerial image understanding and reasoning.

\subsection{Context-Aware Iterative Refinement (CAIR)}
In aerial image QA, tool feedback (e.g., from object detectors) is inherently noisy due to occlusion, degraded image quality, and the small scale of targets in overhead imagery. To prevent such errors from corrupting the accumulated reasoning state, we propose the \textbf{Context-Aware Iterative Refinement (CAIR)} strategy. CAIR augments the standard ReAct loop with a step-level verification stage: after each tool call, a dedicated agent assesses whether the new evidence is reliable and consistent, and selectively updates the answer and accumulated key clues accordingly.

\textbf{Global Reasoning Track.} The \textbf{Strategic Reasoning Agent ($\text{Agent}_{\text{SR}}$)} maintains a ReAct~\cite{yao2022react} loop over the dialogue history $\mathcal{H}_{t-1}$ of the current trace, selecting actions $A_t$ from the tool set provided by DSPE and accumulating tool feedback $F_t$:
\begin{equation}
  \label{eq2}
  \mathcal{H}_t = \mathcal{H}_{t-1} || \{A_t, F_t\}.
\end{equation}

\textbf{Step-level Verification.} At each reasoning step, the \textbf{Perceptual Verification Agent ($\text{Agent}_{\text{PV}}$)} receives only the current step's data: image $I$, query $Q$, action $A_t$, and tool feedback $F_t$, without access to the accumulated history. This isolation prevents historical anchoring bias and produces a draft answer $\hat{a}_t$ and a compact evidence summary $\hat{C}_t$. The \textbf{Contextual Integration Agent ($\text{Agent}_{\text{CI}}$)} does not receive the raw tool output $F_t$; instead, it compares $(\hat{a}_t,\hat{C}_t)$ with the original image--query context and the previous state $(a_{t-1},C_{t-1})$. The image $I$ and query $Q$ are persistent shared context and are omitted from repeated agent-call notation for brevity. The resulting state update is
\begin{equation}
  \label{eq3}
  (a_t,\, C_t) =
  \begin{cases}
    (\hat{a}_t,\;\hat{C}_t) & \text{if trusted and } \hat{a}_t \neq a_{t-1}, \\
    (a_{t-1},\; C_{t-1} || \hat{C}_t) & \text{if trusted and } \hat{a}_t = a_{t-1}, \\
    (a_{t-1},\; C_{t-1}) & \text{otherwise,}
  \end{cases}
\end{equation}

\noindent where $||$ denotes text-level concatenation, $a_t$ is the current best answer, and $C_t$ is a structured text record of accumulated key clues. Because the extracted clues are tightly coupled with the specific answer hypothesis, the state is fully replaced when the draft answer changes to avoid evidence contamination between conflicting hypotheses; when it remains consistent, new clues are safely appended to accumulate evidence.

\textbf{Answer Arbitration.} Upon completion of the reasoning loop, the \textbf{Long-Chain Answer Agent ($\text{Agent}_{\text{LA}}$)} synthesizes an answer $a_{\text{end}}$ and a clue summary $C_{\text{end}}$ from the accumulated history $\mathcal{H}_T$ of that trace. The \textbf{Contextual Integration Agent ($\text{Agent}_{\text{CI}}$)} then selects the more reliable result between $a_{\text{end}}$ and the final iterative state $a_T$ as the trace output $a_{\text{trace}}$.

CAIR's step-level verification protects the answer--evidence state within each trace. During DAAS, every child receives a copy of its parent's state and maintains an independent history, so feedback on an unreliable or pruned branch never enters a sibling branch. The complete procedure is summarized in Algorithm~\ref{alg:cair}.

\begin{algorithm}[!t]
  \caption{Context-Aware Iterative Refinement (CAIR)}
  \label{alg:cair}
\begin{algorithmic}[1]
  \STATE {\bfseries Input:} Task Query $Q$, Image $I$, Iteration Limit $T$
  \STATE {\bfseries Output:} Single Trace Final Answer $a_{\text{trace}}$
  \STATE {\bfseries Initialize:} $\mathcal{H}_0 \leftarrow \{Q, I\}$, $a_0 \leftarrow \text{VLM}(Q, I)$
  \STATE \quad $C_0 \leftarrow \emptyset$, $\mathcal{T}_{\text{opt}} \leftarrow \text{ToolSelection}(Q, I, \mathcal{T})$
  \FOR{$t=1$ {\bfseries to} $T$}
    \STATE \textbf{\textit{Phase 1: Global Reasoning (ReAct)}}
    \STATE \hspace{1em} $A_t \leftarrow \text{Agent}_{\text{SR}}(\mathcal{H}_{t-1}, \mathcal{T}_{\text{opt}})$
    \STATE \hspace{1em} $F_t \leftarrow \text{Execute}(A_t)$
    \STATE \hspace{1em}  Get $\mathcal{H}_t$ based on Eq.~\ref{eq2}
    \STATE \textbf{\textit{Phase 2: Step-level Verification}}
    \STATE \hspace{1em} $\hat{a}_t, \hat{C}_t \leftarrow \text{Agent}_{\text{PV}}(I, Q, A_t, F_t)$
    \STATE \hspace{1em} $(a_t, C_t) \leftarrow \text{Agent}_{\text{CI}}(\hat{a}_t, \hat{C}_t, a_{t-1}, C_{t-1})$ \textit{via Eq.~\ref{eq3}}
  \ENDFOR
  \STATE \textbf{\textit{Phase 3: Answer Arbitration}}
  \STATE $a_{\text{end}}, C_{\text{end}} \leftarrow \text{Agent}_{\text{LA}}(\mathcal{H}_T)$
  \STATE $a_{\text{trace}} \leftarrow \text{Agent}_{\text{CI}}(a_{\text{end}}, C_{\text{end}}, a_T, C_T)$
  \STATE {\bfseries Return} $a_{\text{trace}}$
\end{algorithmic}
\end{algorithm}
\begin{table*}[!t]
\centering
\caption{Comparison with other methods on our UAVQA-Bench. Gen. and Spec. denote the use of the original general-purpose toolkit and our domain-specific toolkit, respectively, both equipped with Qwen3-VL as the backbone. OA and AA denote Overall Accuracy and Average Accuracy, respectively. The GLM4.6V result under UAV-MAS is a supplemental backbone test. Human Avg. reports the average performance of 10 participants. All values are reported in percentages (\%). \textbf{Bold} indicates the best model performance, excluding the human baseline.}
\label{tab:main_results}
\small
\setlength{\tabcolsep}{4pt}
\begin{tabular*}{\textwidth}{@{\extracolsep{\fill}}lllcccccccc}
\toprule
\textbf{Category} & \textbf{Method} & \textbf{Model} & \textbf{OA} & \textbf{AA} & \textbf{ED} & \textbf{CR} & \textbf{QA} & \textbf{FAP} & \textbf{SRU} & \textbf{VG} \\
\midrule

\multicolumn{3}{l}{\textit{Human Avg. (10 participants)}} & 87.87 & 86.14 & 84.33 & 90.40 & 85.43 & 92.50 & 80.80 & 83.39 \\
\midrule
\multirow{10}{*}{\textit{\shortstack{Open Source \\ VLM}}} & \multirow{6}{*}{\textit{Qwen3-VL~\cite{Qwen3-VL}}}
& Qwen3-VL 8B Instruct & 61.73 & 59.63 & 72.33 & 52.80 & 64.57 & 46.50 & 41.60 & 80.00 \\
& & Qwen3-VL 8B Thinking & 52.60 & 50.88 & 79.00 & 50.40 & 64.86 & 44.50 & 39.60 & 26.91 \\
& & Qwen3-VL 30BA3B Instruct & 61.13 & 59.72 & 75.00 & 54.40 & 58.57 & 51.50 & 43.20 & 75.64 \\
& & Qwen3-VL 30BA3B Thinking & 62.00 & 60.03 & 78.33 & 52.00 & 69.43 & 53.50 & 56.00 & 50.91 \\
& & Qwen3-VL 32B Instruct & 69.60 & 68.63 & 80.00 & 67.20 & 67.71 & 58.00 & 59.60 & 79.27 \\
& & Qwen3-VL 32B Thinking & 68.27 & 67.06 & 77.33 & 63.20 & 67.14 & 57.00 & 58.40 & 79.27 \\
\cmidrule{2-11}
& \multirow{2}{*}{\textit{InternVL3.5~\cite{wang2025internvl3}}} 
&InternVL3.5 8B & 40.53 & 39.18 & 58.00 & 36.00 & 53.71 & 36.50 & 47.20 & 3.63 \\
& & InternVL3.5 38B & 55.13 & 55.73 & 77.33 & 66.40 & 61.14 & 60.50 & 51.20 & 17.82 \\
\cmidrule{2-11}
& \textit{GLM4.6V~\cite{vteam2025glm45vglm41vthinkingversatilemultimodal}} 
& GLM4.6V 9B & 65.27 & 63.57 & 72.00 & 57.60 & 67.71 & 51.00 & 56.40 & 76.73 \\
\cmidrule{2-11}
& 
\textit{Qwen3.5~\cite{qwen35blog}}
& Qwen3.5 9B & 63.80 & 62.81 & 65.67 & 59.20 & 65.14 & 57.5 & 50.80 & 78.55 \\
\midrule
\multirow{4}{*}{\textit{\shortstack{Closed Source \\ VLM}}} & \multirow{2}{*}{\textit{ChatGPT~\cite{openai_gpt5_2}}} 
&ChatGPT 5.2 Pro &57.13 &56.72 &77.67 &60.80 &71.43 &58.50 &67.20 &4.73 \\
& &ChatGPT 5.2 & 53.20&52.30 &75.67 &57.60 & 69.43& 46.50 &58.80 & 5.82\\
\cmidrule{2-11}
& \multirow{2}{*}{\textit{Gemini~\cite{google_gemini_2026}}} 
&Gemini 3 Pro &73.00 &71.25 &79.67 &60.80 &76.29 &68.50 &60.80 &81.45 \\
& &Gemini 3 Flash &70.73 &68.78 &\textbf{82.33} &59.20 &71.71 &60.50 &56.40 & 82.55\\

\midrule
\multirow{10}{*}{\textit{\shortstack{Agent Based \\ Method}}} & \multirow{2}{*}{\textit{DyFo~\cite{li2025dyfo}}} 
& Qwen3-VL 8B &46.47 & 48.31 & 61.00 & 64.00 & 48.86 & 63.00 & 35.20 & 17.82 \\
& & Qwen3-VL 32B & 51.67 & 52.81 &  70.00 & 64.00 & 57.14 & 61.50 & 58.40 & 5.82 \\
\cmidrule{2-11}
& \multirow{4}{*}{\textit{Qwen-Agent~\cite{qwen_agent}}} 
& Qwen3-VL 8B-Gen. & 60.87 & 59.11 & 75.67 & 52.00 &61.43& 52.00 & 41.20 & 72.36 \\
& & Qwen3-VL 8B-Spec. &61.13 &  60.46 & 71.00	& 60.00 & 	56.57 &  50.50 & 51.60 & 73.09\\
& & Qwen3-VL 32B-Gen. &71.13 &  71.33 & 74.33 & 78.40 &  77.43 &  72.50 &  58.40 & 66.91 \\
& & Qwen3-VL 32B-Spec. & 69.40 & 69.63 & 63.67 & 71.20 &  71.43 & 71.50 & 68.00 & 72.00 \\
\cmidrule{2-11}
& \multirow{4}{*}{\textit{PyVision~\cite{zhao2025pyvision}}} 
& Qwen3-VL 8B-Gen. & 44.80 & 44.20 & 64.00 & 48.80 & 63.14 & 51.50 & 35.20 &	2.55 \\
& & Qwen3-VL 8B-Spec. &50.87 & 49.16 & 62.67 & 45.60 & 63.71 & 44.50 & 39.20 & 39.27 \\
& & Qwen3-VL 32B-Gen. & 50.60 & 51.74 & 63.67 & 68.00 & 69.71 & 67.50 & 41.20 &	0.36 \\
& & Qwen3-VL 32B-Spec. & 52.33 & 52.53 & 62.67 & 58.40 &  61.43 & 59.00 & 46.40 & 27.27 \\
\midrule
\multirow{4}{*}{\textbf{Ours}} & \multirow{4}{*}{\textit{UAV-MAS}} & \textbf{Qwen3-VL 8B} & 70.47 & 68.94 & 77.67 & 61.60 & 69.71 & 57.50 & 66.80 & 80.36 \\
& & \textbf{Qwen3-VL 32B} & \textbf{77.00}&\textbf{76.03} &77.00 &\textbf{70.40}& \textbf{80.57}&\textbf{75.00} & 69.20&\textbf{84.00} \\
& & \textbf{GLM4.6V 9B} & 69.67 & 67.67 & 75.67 & 56.80 & 70.86 & 54.00 & \textbf{71.60} & 77.09 \\
& & \textbf{Qwen3.5 9B} & 73.67 & 73.81 & 74.00 & 70.40 & 74.57 & 70.50 & 67.20 & 81.81 \\

\bottomrule
\end{tabular*}
\end{table*}
\subsection{Difficulty-Aware Adaptive Search (DAAS)}
While CAIR mitigates noise at each reasoning step, it follows a fixed linear chain that may miss better reasoning paths for complex aerial queries. In UAV visual question answering, query difficulty varies substantially: a simple presence check requires few reasoning steps, while a compositional query (e.g., counting objects satisfying multiple spatial and attribute conditions) demands deeper, multi-step exploration. Applying a fixed pruning threshold across all queries wastes computation on easy queries and prematurely cuts off exploration on hard ones. To address this, we propose the \textbf{Difficulty-Aware Adaptive Search (DAAS)} strategy. Unlike standard beam search, which applies a single fixed pruning threshold and selects the final-node answer, DAAS derives a per-query pruning threshold from estimated query difficulty and selects the optimal path by global coherence across all nodes. DAAS governs tree expansion through three phases: Adaptive Initialization, Consistency-Guided Branch Control, and Optimal Path Selection.

\textbf{Adaptive Initialization.} Before searching, the \textbf{Score Agent ($\text{Agent}_{\text{SC}}$)} assigns the base model's direct response an initial score $S_{\text{init}}\in[0,10]$ as a coarse difficulty indicator. We use a fixed mapping: $S_{\text{init}}\in[0,3]$, $[4,8]$, and $[9,10]$ correspond to $\tau=2$, $4$, and $6$, respectively. Larger thresholds avoid unnecessary expansion for high-confidence queries, while smaller thresholds preserve exploration for low-confidence queries.

\textbf{Consistency-Guided Branch Control.} We manage tree expansion through a Consecutive Consistency check. During expansion, the system generates $W$ candidate successor states via CAIR using temperature-based sampling, where $W$ controls branching width. The \textbf{Score Agent ($\text{Agent}_{\text{SC}}$)} evaluates each candidate's quality to output a new score $S'$. Branch continuation is decided by coherence over consecutive steps:
%\begin{itemize}[itemsep=1pt, topsep=2pt, parsep=0pt, partopsep=0pt]
\begin{itemize}
    \item \textbf{Prune Branch:} We prune only if both the current node and the new candidate fall below the threshold simultaneously:
    \begin{equation}
    \label{eq5}
    \text{Prune} \iff (S' < \tau) \land (S < \tau).
    \end{equation}
    
    This allows a single low-confidence step to persist if its adjacent step remains high-confidence. The key insight is that intermediate tool calls in aerial reasoning (e.g., a zoom step returning a blurry sub-patch before a subsequent description step integrates the result) may transiently score low without indicating a dead branch. Requiring two \emph{consecutive} low-confidence steps before pruning reduces false terminations while discarding genuinely unproductive paths.
    \item \textbf{Expand Branch:} If at least one step in the consecutive pair maintains high confidence, we create a new node $N'$ and continue exploration from it.
\end{itemize}

\textbf{Optimal Path Selection.} Tree expansion follows branch-local recursion and terminates when a valid final answer is generated or the maximum depth is reached. A final answer is valid only if it matches the required task format, namely a candidate option for multiple-choice tasks or a parseable normalized box for visual grounding. Paths that reach the depth limit without a valid answer are discarded. Upon completion, we select the optimal valid path by evaluating global coherence across all nodes rather than relying solely on the final-node score. To mitigate path-length bias commonly associated with average scoring across long chains, $\text{Agent}_{\text{SC}}$ is explicitly prompted to assign strict penalties to uninformative or meandering intermediate steps. The optimal path $\mathcal{P}^*$ is thus selected by maximizing the average node score:
\begin{equation}
  \label{eq6}
\mathcal{P}^* = \arg\max_{\mathcal{P}_i} \left( \frac{1}{|\mathcal{P}_i|} \sum_{N \in \mathcal{P}_i} S(N) \right).
\end{equation}

Subsequently, the valid answer stored at the leaf of $\mathcal{P}^*$ is selected as the global final answer $a_{\text{final}}$:
\begin{equation}
  \label{eq7}
a_{\text{final}} \leftarrow a_{\text{leaf}}(\mathcal{P}^*).
\end{equation}

If every branch is pruned or reaches the depth limit without a valid answer, DAAS returns the base model's initial response $a_0$ as a conservative fallback. In summary, DAAS focuses computation on aerial queries that require multi-step spatial or attribute reasoning by deriving a per-query pruning threshold and selecting the valid path with the highest global coherence. The complete procedure is summarized in Algorithm~\ref{alg:daas}. Implementation details and agent configurations are provided in the supplementary material.

\begin{algorithm}[!t]
   \caption{Difficulty-Aware Adaptive Search (DAAS)}
   \label{alg:daas}
\begin{algorithmic}[1]
   \STATE {\bfseries Input:} Image $I$, Query $Q$, Max Depth $D$, Max Width $W$
   \STATE {\bfseries Output:} Final Answer $a_{\text{final}}$
   \STATE $a_0 \leftarrow \text{VLM}(Q,I)$, $S_{\text{init}} \leftarrow \text{Agent}_{\text{SC}}(Q,I,a_0)$
   \STATE $\tau \leftarrow g(S_{\text{init}})$ using the fixed three-level map
   \STATE $N_{\text{root}} \leftarrow \{\mathcal{H}_0,a_0,C_0,S_{\text{init}},d=0\}$, $\mathcal{V}\leftarrow\emptyset$
   \STATE \textbf{Function Search}($N\{\mathcal{H},a,C,S,d\}$)
   \FOR{$k=1$ {\bfseries to} $W$}
      \STATE Sample one CAIR step to obtain $\mathcal{H}',a',C'$
      \STATE $S' \leftarrow \text{Agent}_{\text{SC}}(\mathcal{H}'-\mathcal{H})$
      \IF{$S'<\tau$ $\land$ $S<\tau$}
         \STATE {\bfseries continue} \textit{// prune two consecutive weak steps}
      \ENDIF
      \STATE $N'\leftarrow\{\mathcal{H}',a',C',S',d+1\}$
      \IF{$\text{Final}(a')$ $\land$ $\text{Valid}(a')$}
         \STATE $\mathcal{V}\leftarrow\mathcal{V}\cup\{\mathcal{P}(N')\}$
      \ELSIF{$d+1=D$}
         \IF{$\text{Valid}(a')$}
            \STATE $\mathcal{V}\leftarrow\mathcal{V}\cup\{\mathcal{P}(N')\}$
         \ENDIF
      \ELSE
         \STATE \textbf{Search}($N'$)
      \ENDIF
   \ENDFOR
   \STATE {\bfseries End Function}
   \STATE \textbf{Search}($N_{\text{root}}$)
   \IF{$\mathcal{V}=\emptyset$}
      \STATE $a_{\text{final}}\leftarrow a_0$ \textit{// fallback for pruned/invalid searches}
   \ELSE
      \STATE Select $\mathcal{P}^*$ from $\mathcal{V}$ using Eq.~\ref{eq6}
      \STATE $a_{\text{final}}\leftarrow a_{\text{leaf}}(\mathcal{P}^*)$ via Eq.~\ref{eq7}
   \ENDIF
   \STATE {\bfseries Return} $a_{\text{final}}$
\end{algorithmic}
\end{algorithm}

\section{Experiment}
\subsection{Main Results}

The quantitative results on UAVQA-Bench are presented in Table~\ref{tab:main_results}. Built upon Qwen3-VL 8B and 32B models, our proposed UAV-MAS demonstrates substantial performance gains over both vanilla baselines and existing agent frameworks. Specifically, UAV-MAS-8B and UAV-MAS-32B surpass their respective Instruct counterparts by margins of 8.7\% and 7.4\% in Overall Accuracy. This substantial improvement underscores the efficacy of our design in bridging the domain gap that limits standard MLLMs. Furthermore, compared to general-purpose multi-agent frameworks utilizing identical models, UAV-MAS achieves a 5.9\% lead over the strongest baseline, validating the effectiveness of our designs. Remarkably, UAV-MAS-32B outperforms the closed-source Gemini 3 Pro by 4.0\%, showing that a domain-specialized open-source multi-agent design can compete effectively with proprietary general-purpose models.

\subsection{Ablation Study}
\subsubsection{Module-Level Ablation}
\begin{table}[!t]
\centering
\caption{Module-level ablation of UAV-MAS on UAVQA-Bench. Baseline: Qwen3-VL 8B Instruct. OA: Overall Accuracy (\%).}
\label{tab:ablation_module}
\small
\begin{tabular}{lcccc}
\toprule
{\textbf{Method}} & {\textbf{DSPE}} & {\textbf{CAIR}} & {\textbf{DAAS}} & {\textbf{OA}} \\
\midrule
Baseline &            &            &            & 61.73 \\
Exp1     & \checkmark &            &            & 64.53 \\
Exp2     & \checkmark & \checkmark &            & 68.03 \\
Exp3     & \checkmark &            & \checkmark & 68.27 \\
Full     & \checkmark & \checkmark & \checkmark & \textbf{70.47} \\
\bottomrule
\end{tabular}
\end{table}

\begin{figure*}[!t]
  \begin{center}
    \includegraphics[width=\textwidth]{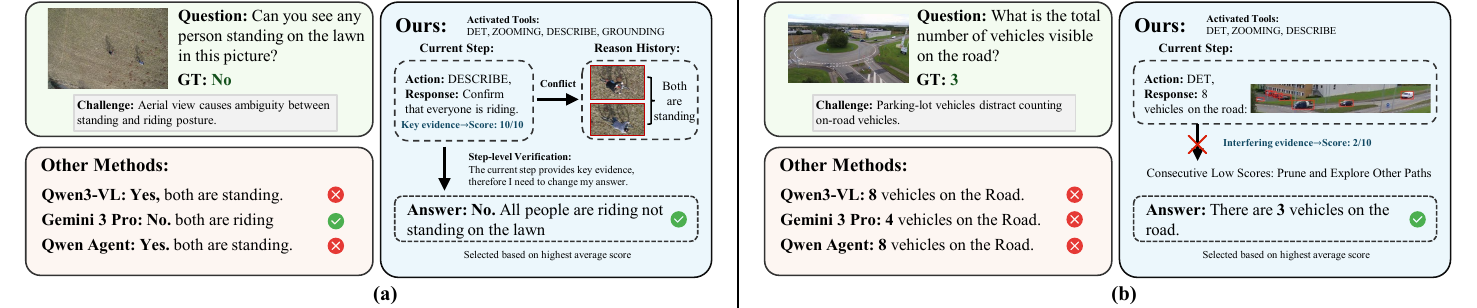}
    \caption{Visual comparison of UAV-MAS with other methods. All experiments are based on Qwen3-VL 8B except Gemini 3 Pro.}
    \label{fig:fig_visual}
  \end{center}
\end{figure*}
Table~\ref{tab:ablation_module} presents an incremental analysis of each module. Starting from the vanilla Qwen3-VL 8B baseline (61.73\%), equipping it with DSPE alone (Exp1) raises accuracy to 64.53\%, confirming the necessity of domain-specific perception operators. When CAIR is further added (Exp2), performance jumps substantially to 68.03\%, reflecting its effectiveness in suppressing error propagation across reasoning steps. Adding DAAS instead of CAIR (Exp3) also yields a strong gain to 68.27\%, demonstrating that adaptive search independently contributes to accurate reasoning. Integrating all three modules achieves the highest score of 70.47\%, indicating that DSPE, CAIR, and DAAS address complementary bottlenecks and together produce a synergistic effect.

\subsubsection{Within-Module Design Ablation}

Table~\ref{tab:ablation_design} examines the internal design choices of each module against the full system (70.47\%).
% For DSPE, we first compare toolkit types in isolation (without Per-Tool Activation): equipping the model with general-purpose tools actually \emph{hurts} performance (60.87\%), as the domain gap between generic tools and aerial tasks introduces noise. Domain-specific tools recover to 61.13\%, validating their necessity. When Per-Tool Activation is disabled in the full system, accuracy drops to 69.60\%, and collapsing the distributed ToolSelection Agents into a single centralized agent causes a larger decline to 68.07\%, confirming that per-tool, independent activation is critical for filtering irrelevant operators.
For DSPE, disabling Per-Tool Activation---which feeds all available tools directly into the agent---causes accuracy to drop to 69.60\%, highlighting the critical importance of actively selecting task-relevant tools. Furthermore, collapsing the parallel, distributed $\text{Agent}_{\text{TS}}$ into a single centralized agent degrades performance to 68.07\%, demonstrating that in domain-specific scenarios, smaller, less capable models require highly fine-grained tool management to operate effectively. For CAIR, completely removing the module yields 68.27\%. Ablating $\text{Agent}_{\text{CI}}$ alone still results in a 0.20\% drop, showing that contextual integration is needed to reconcile verification signals; ablating $\text{Agent}_{\text{PV}}$ causes a more pronounced decline to 69.53\%, highlighting that independent perceptual verification is the more critical of the two agents. For DAAS, removing the tree-search mechanism entirely (w/o DAAS) yields 68.03\%. Disabling difficulty awareness (69.87\%) or the pruning mechanism (68.73\%) also incurs clear penalties, demonstrating that both adaptive resource allocation and branch filtering are essential for high-quality search.

\begin{table}[!t]
\centering
\caption{Within-module design ablation of UAV-MAS. Baseline: Qwen3-VL 8B Instruct. OA: Overall Accuracy (\%).}
\label{tab:ablation_design}
\small
\begin{tabular}{l l c}
\toprule
\textbf{Module} & \textbf{Variant} & \textbf{OA (\%)} \\
\midrule
Baseline & - & 61.73 \\
\midrule
\multirow{2}{*}{\textit{DSPE}}
  & w/o Per-Tool Activation           & 69.60 \\
  & w/o Distributed $\text{Agent}_{\text{TS}}$          & 68.07 \\
\midrule
\multirow{3}{*}{\textit{CAIR}}
  & w/o CAIR                          & 68.27 \\
  & w/o $\text{Agent}_{\text{CI}}$    & 70.27 \\
  & w/o $\text{Agent}_{\text{PV}}$    & 69.53 \\
\midrule
\multirow{3}{*}{\textit{DAAS}}
  & w/o DAAS                          & 68.03 \\
  & w/o Difficulty Awareness          & 69.87 \\
  & w/o Pruning                       & 68.73 \\
\midrule
\textbf{Full}&-           & \textbf{70.47} \\
\bottomrule
\end{tabular}
\end{table}

\subsubsection{DAAS Efficiency Ablation}
To assess the computational role of DAAS, Table~\ref{tab:efficiency} reports average per-query results on one NVIDIA H200 GPU. Compared with UAV-MAS without DAAS, the full method introduces additional search computation while improving OA from 68.03\% to 70.47\%. More importantly, at a similar accuracy level, Majority Vote@3 applied to UAV-MAS without DAAS reaches 70.73\% OA but requires 265.29~s, 36.39 MLLM calls, and 5.07 tool calls per query. Full UAV-MAS obtains 70.47\% OA using 112.23~s, 25.60 MLLM calls, and 2.99 tool calls, reducing latency by 57.7\% while requiring fewer model and tool calls. These results show that DAAS provides a better accuracy--cost trade-off than brute-force repeated sampling through targeted exploration.

\begin{table}[!h]
\centering
\caption{Efficiency ablation of DAAS on one NVIDIA H200 GPU. MV3 denotes Majority Vote@3.}
\label{tab:efficiency}
\scriptsize
\setlength{\tabcolsep}{3.0pt}
\begin{tabular}{@{}lrrrr@{}}
\toprule
\textbf{Method} & \textbf{OA (\%)} & \textbf{Time (s)} & \textbf{MLLM Calls} & \textbf{Tool Calls} \\
\midrule
UAV-MAS w/o DAAS & 68.03 & 88.43 & 12.13 & 1.69 \\
UAV-MAS w/o DAAS + MV3 & 70.73 & 265.29 & 36.39 & 5.07 \\
\textbf{UAV-MAS} & \textbf{70.47} & \textbf{112.23} & \textbf{25.60} & \textbf{2.99} \\
\bottomrule
\end{tabular}
\end{table}

\subsection{Visual Comparison}
Figure~\ref{fig:fig_visual} presents two representative cases where baselines fail and our UAV-MAS succeeds, collectively demonstrating all three modules. In both cases, DSPE's per-tool activation first selects a task-relevant tool subset rather than activating all available tools indiscriminately. In Figure~\ref{fig:fig_visual}(a), the extreme top-down viewpoint creates pose ambiguity: the shown open-source baselines misidentify cyclists as persons standing on the lawn. The DESCRIBE tool returns a high-confidence observation that conflicts with the prior belief; CAIR's step-level verification agent detects this conflict and updates the reasoning state accordingly, yielding the correct answer. This demonstrates that step-level verification is essential for correcting perceptual bias before it propagates through the reasoning chain. In Figure~\ref{fig:fig_visual}(b), the challenge is semantic distraction, manifesting in two distinct failure modes: the shown open-source baselines over-count to 8 by including parking-lot vehicles, while Gemini 3 Pro hallucinates a non-existent on-road vehicle and returns 4. DAAS assigns a low confidence score to the distractor-contaminated detection; two consecutive low scores then trigger branch pruning, and the search continues along alternative paths that correctly isolate the 3 on-road vehicles. This demonstrates that consecutive-confidence pruning effectively rejects misleading evidence and selects the globally more reliable reasoning trace.

\subsection{Cross-Dataset Generalization on CHOICE}
To evaluate cross-dataset generalization, we test UAV-MAS-8B on the independent CHOICE remote-sensing benchmark~\cite{an2026choice}, using only generic image zooming and description tools, and compare it with the Qwen3-VL-8B Instruct and Thinking variants. We retain 440 questions whose answer formats are compatible with our evaluation and exclude Referring Expression Segmentation (RES) from Cross-instance Discernment because it requires pixel-level masks. Following the CHOICE taxonomy, ILC, SII, CID, AttR, AssR, and CSR denote Image-level Comprehension, Single-instance Identification, Cross-instance Discernment, Attribute Reasoning, Assessment Reasoning, and Common Sense Reasoning, respectively. As shown in Table~\ref{tab:choice_generalization}, UAV-MAS-8B achieves the highest Overall Accuracy of 75.23\%, outperforming the two baselines by 3.41 and 6.14 percentage points, respectively. These results demonstrate the effectiveness and generalizability of our method beyond the UAVQA-Bench distribution and UAV-specific perception tools.

\begin{table}[!t]
\centering
\caption{Cross-dataset generalization on the CHOICE subset. All values are percentages (\%).}
\label{tab:choice_generalization}
\scriptsize
\setlength{\tabcolsep}{2.4pt}
\begin{tabular}{@{}lccccccc@{}}
\toprule
\textbf{Method} & \textbf{OA} & \textbf{ILC} & \textbf{SII} & \textbf{CID} & \textbf{AttR} & \textbf{AssR} & \textbf{CSR} \\
\midrule
Qwen3-VL-8B Inst. & 71.82 & 75.00 & 72.14 & 75.00 & \textbf{65.00} & 42.50 & 86.67 \\
Qwen3-VL-8B Think. & 69.09 & 70.00 & 68.57 & \textbf{83.33} & 47.50 & 47.50 & 78.33 \\
UAV-MAS-8B & \textbf{75.23} & \textbf{85.00} & \textbf{77.14} & 76.67 & \textbf{65.00} & \textbf{50.00} & \textbf{91.67} \\
\bottomrule
\end{tabular}
\end{table}

\section{Limitations and Future Work}

Despite its encouraging performance, UAV-MAS has two main limitations. First, its iterative multi-agent reasoning and tool invocation introduce non-negligible inference latency. Although employing lightweight models and the DAAS strategy reduces unnecessary computation, the system remains substantially slower than a single forward pass. Consequently, it is currently more suitable for offline image analysis at a ground station than for real-time on-board inference. Second, its performance still has room for improvement. CAIR mitigates the propagation of local errors during iterative reasoning but cannot eliminate them entirely. In some cases, the model changes a correct answer to an incorrect one, fails to recognize critical visual information, or cannot objectively estimate the benefit of each reasoning step.

Future work will address these limitations from both efficiency and performance perspectives. For efficient deployment, we will investigate parallel agent execution, reusable visual-feature caching, more aggressive early-exit and dynamic-routing mechanisms, and model compression to reduce redundant computation and enable real-time on-board processing. To improve performance, we will explore stronger fine-grained visual perception, uncertainty-aware answer preservation and rollback mechanisms, and better-calibrated process-level verification for estimating step-wise reasoning gains. These directions are expected to make UAV-MAS faster, more reliable, and more practical for real-world UAV applications.

\section{Conclusion}

In this paper, we introduce UAVQA-Bench, a comprehensive, fully human-annotated benchmark for UAV aerial image understanding and reasoning. It covers 6 capability dimensions across 16 tasks (in multiple-choice and visual grounding formats), with 1,500 samples drawn from 13 diverse public UAV datasets. We systematically evaluate a broad range of open-source and closed-source MLLMs as well as agent-based systems on UAVQA-Bench. Based on the results, we present UAV-MAS, a novel training-free multi-agent system designed to address the unique challenges of visual perception and reasoning in UAV scenarios. By integrating a Domain-Specific Perception Engine with advanced reasoning strategies---Context-Aware Iterative Refinement and Difficulty-Aware Adaptive Search---our system effectively overcomes the limitations of existing specialized models and general-purpose agents. Experiments demonstrate that UAV-MAS achieves state-of-the-art performance, enabling a 32B-parameter open-source model to surpass Gemini 3 Pro on UAVQA-Bench. These results underscore the potential of specialized multi-agent architectures in domain-specific applications. Future work will focus on optimizing real-time on-board inference and expanding the system's capabilities to dynamic video understanding tasks.

\bibliographystyle{IEEEtran}
\bibliography{sample-base}

\begin{IEEEbiography}[{\includegraphics[width=1in,height=1.25in,clip,keepaspectratio]{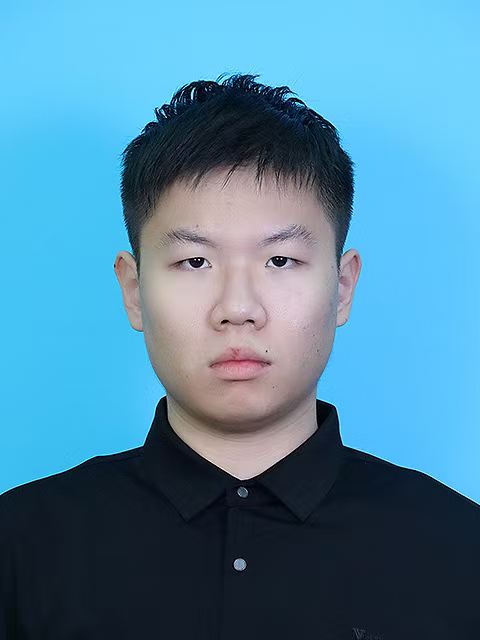}}]{Haoyu Zhang}
received the B.S. degree in electronic science and technology from the School of Information Science and Technology, Fudan University, Shanghai, China, in 2025.
He is currently pursuing the Ph.D. degree with the College of Future Information Technology, Fudan University.
His research interests include computer vision, multimodal large models, and intelligent agent systems.
\end{IEEEbiography}

\begin{IEEEbiography}[{\includegraphics[width=1in,height=1.25in,clip,keepaspectratio]{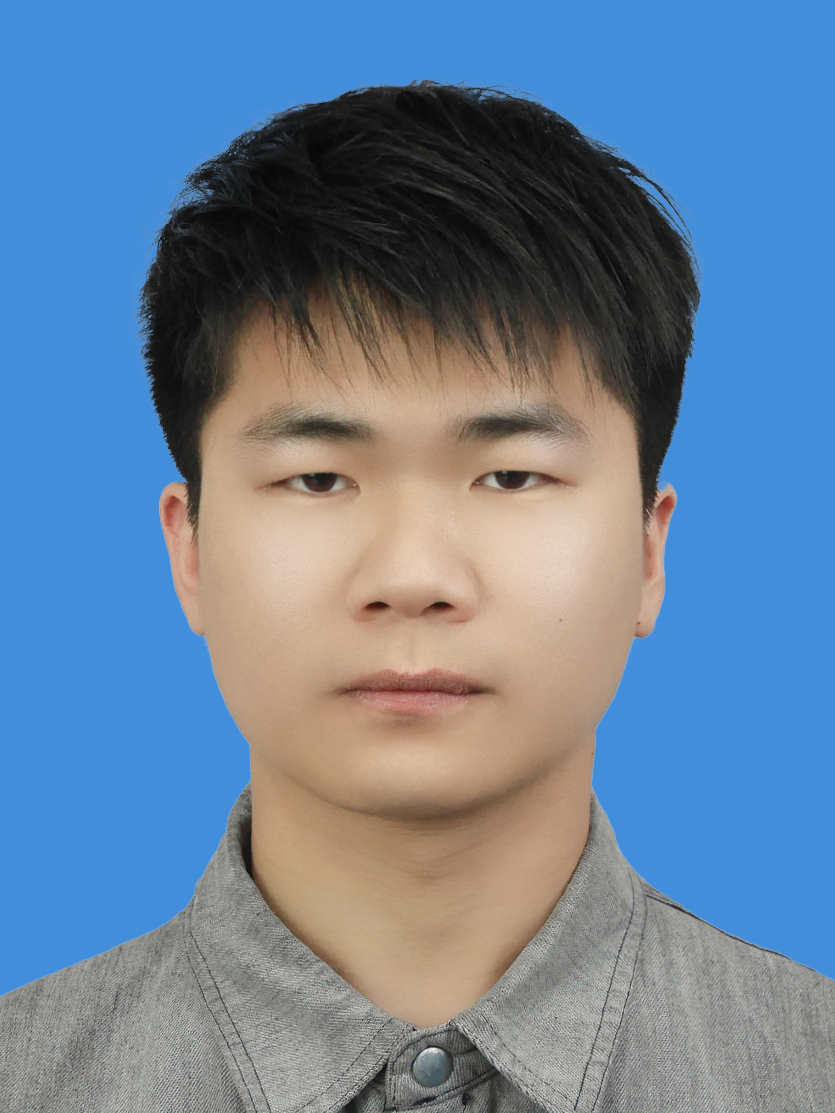}}]{Shuoxun Zhang}
received the B.S. degree in Communication Engineering from Shanghai University, Shanghai, China, in 2023. He is currently pursuing a master's degree at the College of Future Information Technology, Fudan University. His main research interests include multimodal large models and intelligent agent systems.
\end{IEEEbiography}

\begin{IEEEbiography}[{\includegraphics[width=1in,height=1.25in,clip,keepaspectratio]{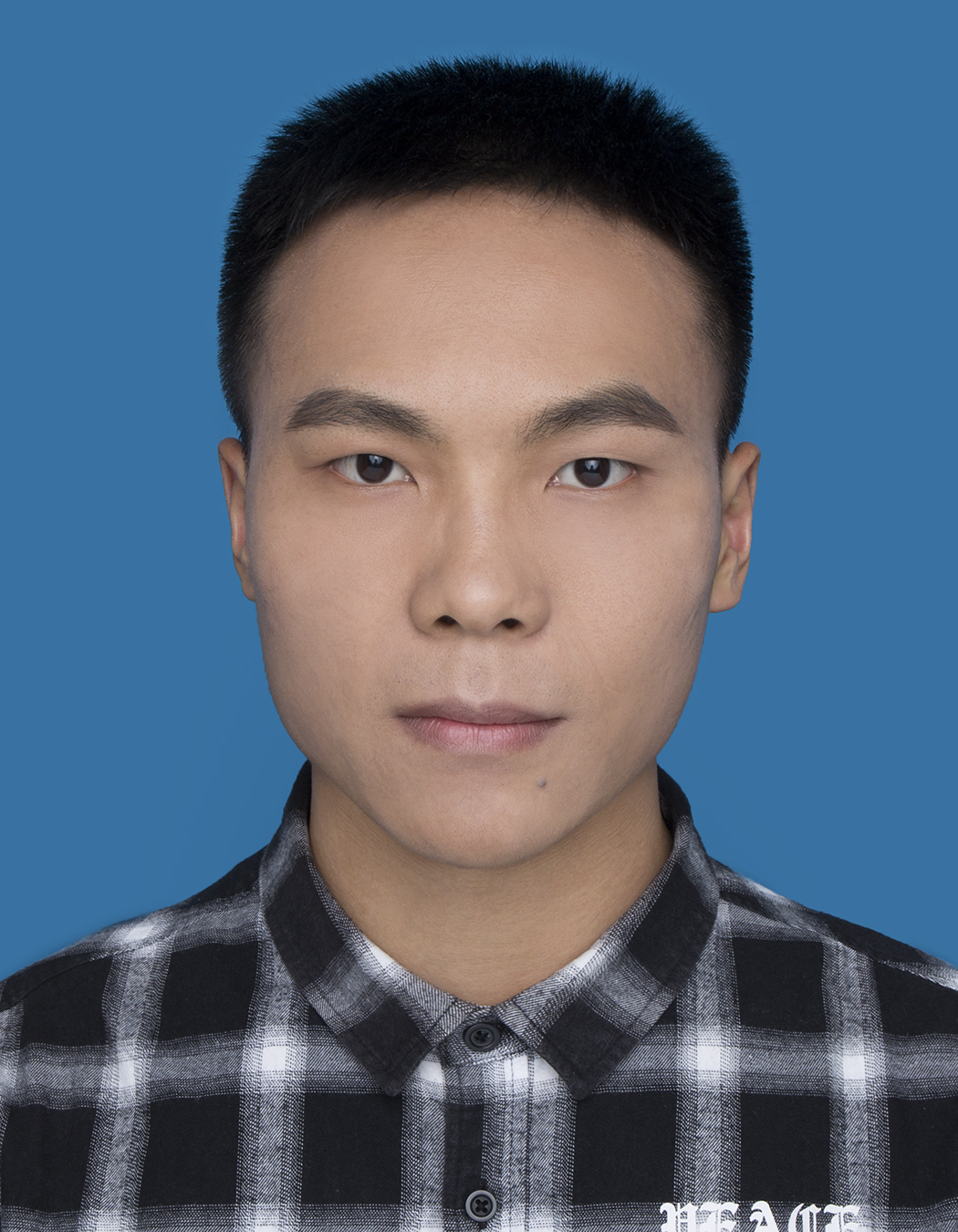}}]{Peng Ye}
is currently a postdoctoral fellow at MMLab of The Chinese University of Hong Kong and a scientific research advisor of Shanghai AI Laboratory. He received the Ph.D. degree from Fudan University, Shanghai, China. His research interests include autonomous agents, (M)LLMs, foundation models, and efficient design and optimization. He has published papers in leading journals and conferences, including IEEE TPAMI, IJCV, CVPR, ICCV, ECCV, NeurIPS, ICML, ACM MM, and ICME. He also serves as a reviewer or program committee member for journals and conferences including IEEE TPAMI, IJCV, CVPR, ECCV, ICCV, and NeurIPS.
\end{IEEEbiography}

\begin{IEEEbiography}[{\includegraphics[width=1in,height=1.25in,clip,keepaspectratio]{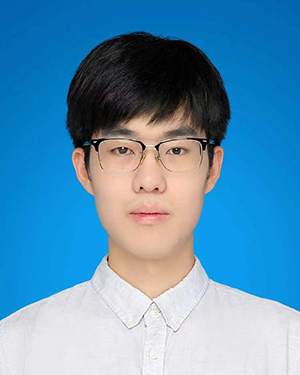}}]{Lin Zhang}
received the B.S. degree in electronic engineering from Fudan University, Shanghai, China, in 2022, where he is currently pursuing the Ph.D. degree with the College of Future Information Technology. His main research interests include computer vision and transfer learning.
\end{IEEEbiography}

\begin{IEEEbiography}[{\includegraphics[width=1in,height=1.25in,clip,keepaspectratio]{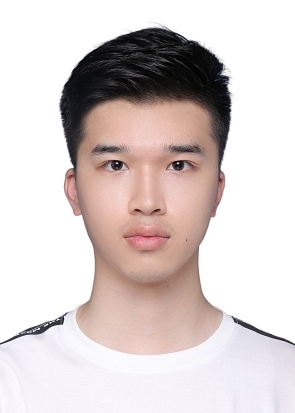}}]{Jiakang Yuan}
is currently pursuing the Ph.D. degree in electronic engineering with the College of Future Information Technology, Fudan University. He received the bachelor's degree in electronic engineering from Fudan University in 2022. His research interests include multimodal reasoning, multi-agent systems, and spatial intelligence. He has published papers in leading journals and conferences, including CVPR, ICCV, ECCV, NeurIPS, and IEEE TPAMI, and serves as a reviewer for journals and conferences including IEEE TIP, IEEE TCSVT, CVPR, ECCV, and ICCV.
\end{IEEEbiography}

\begin{IEEEbiography}[{\includegraphics[width=1in,height=1.25in,clip,keepaspectratio]{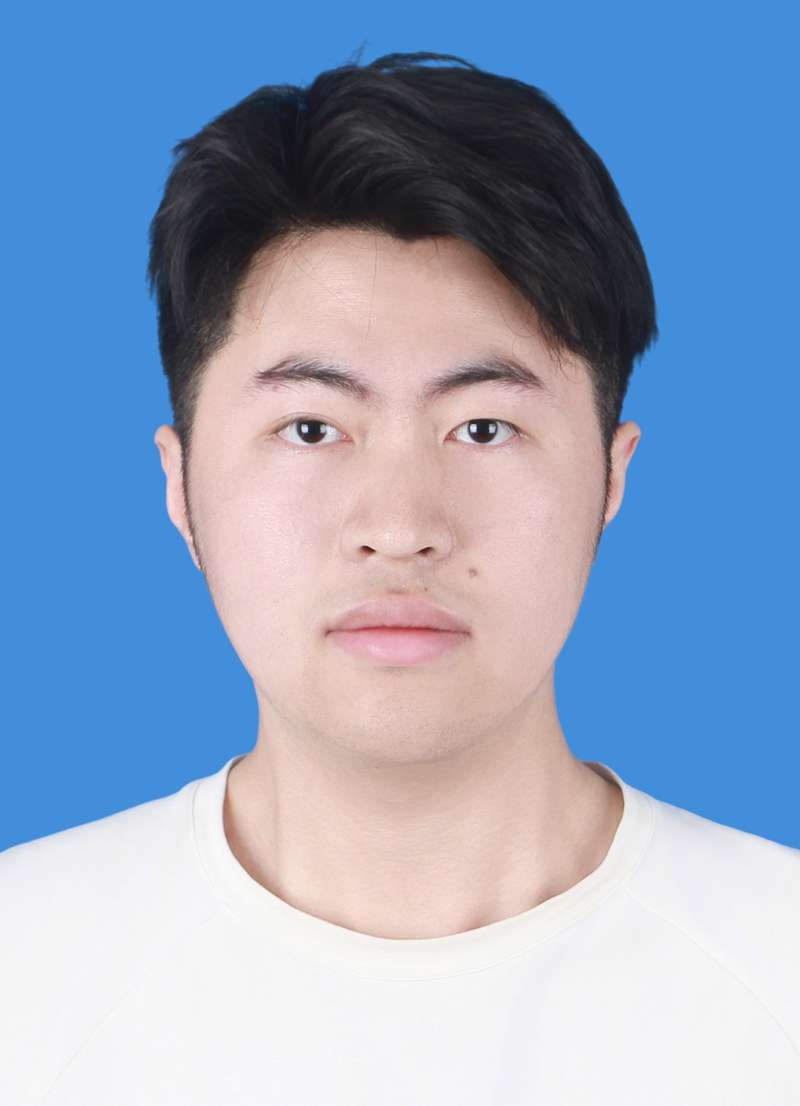}}]{Shenghong Yi}
received the B.S. degree in Intelligent Science and Technology from Fudan University, Shanghai, China, in 2026. He is currently working toward the Ph.D. degree at the College of Future Information Technology. His research interests include computer vision and multimodal large models.
\end{IEEEbiography}

\begin{IEEEbiography}[{\includegraphics[width=1in,height=1.25in,clip,keepaspectratio]{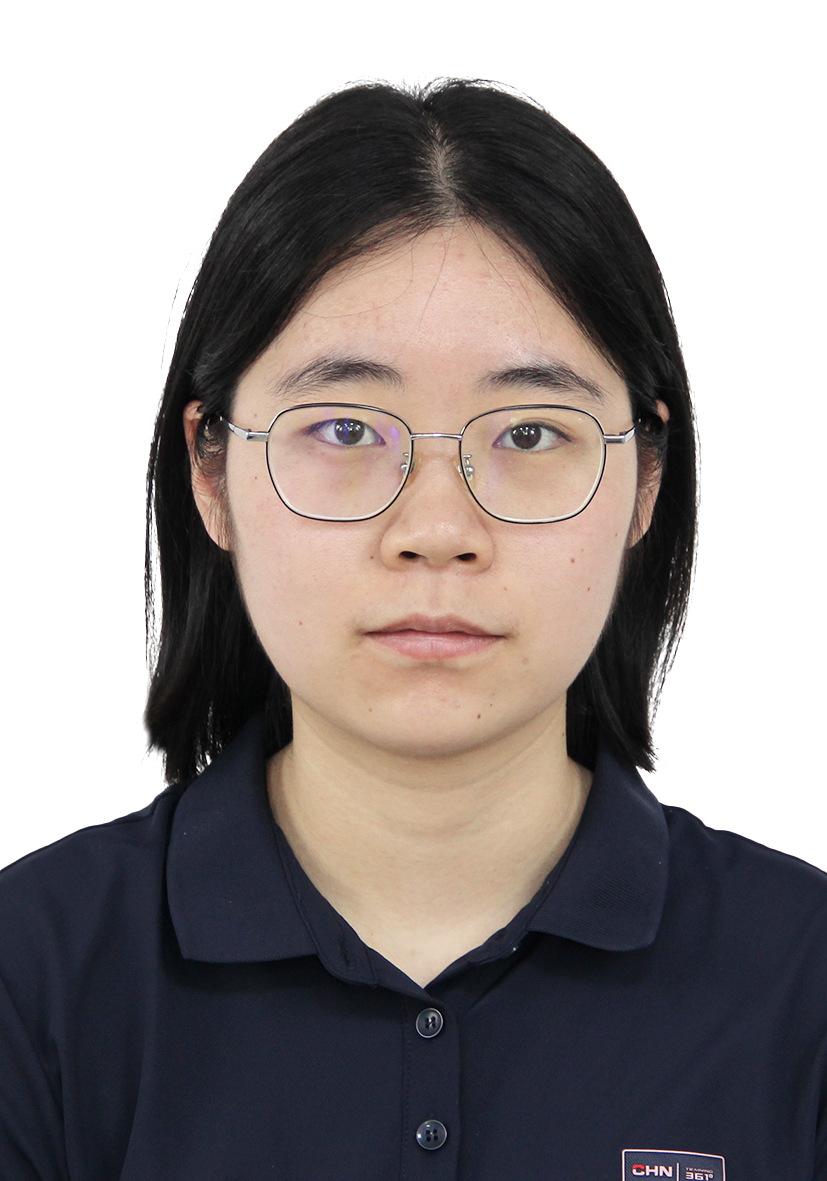}}]{Yuening Wang}
is currently pursuing the bachelor's degree in Electronic Information Science and Technology at the College of Future Information Innovation, Fudan University, Shanghai, China, and is expected to graduate in June 2027.
\end{IEEEbiography}
% TODO: Add the biography and portrait of Yuening Wang.

\begin{IEEEbiography}[{\includegraphics[width=1in,height=1.25in,clip,keepaspectratio]{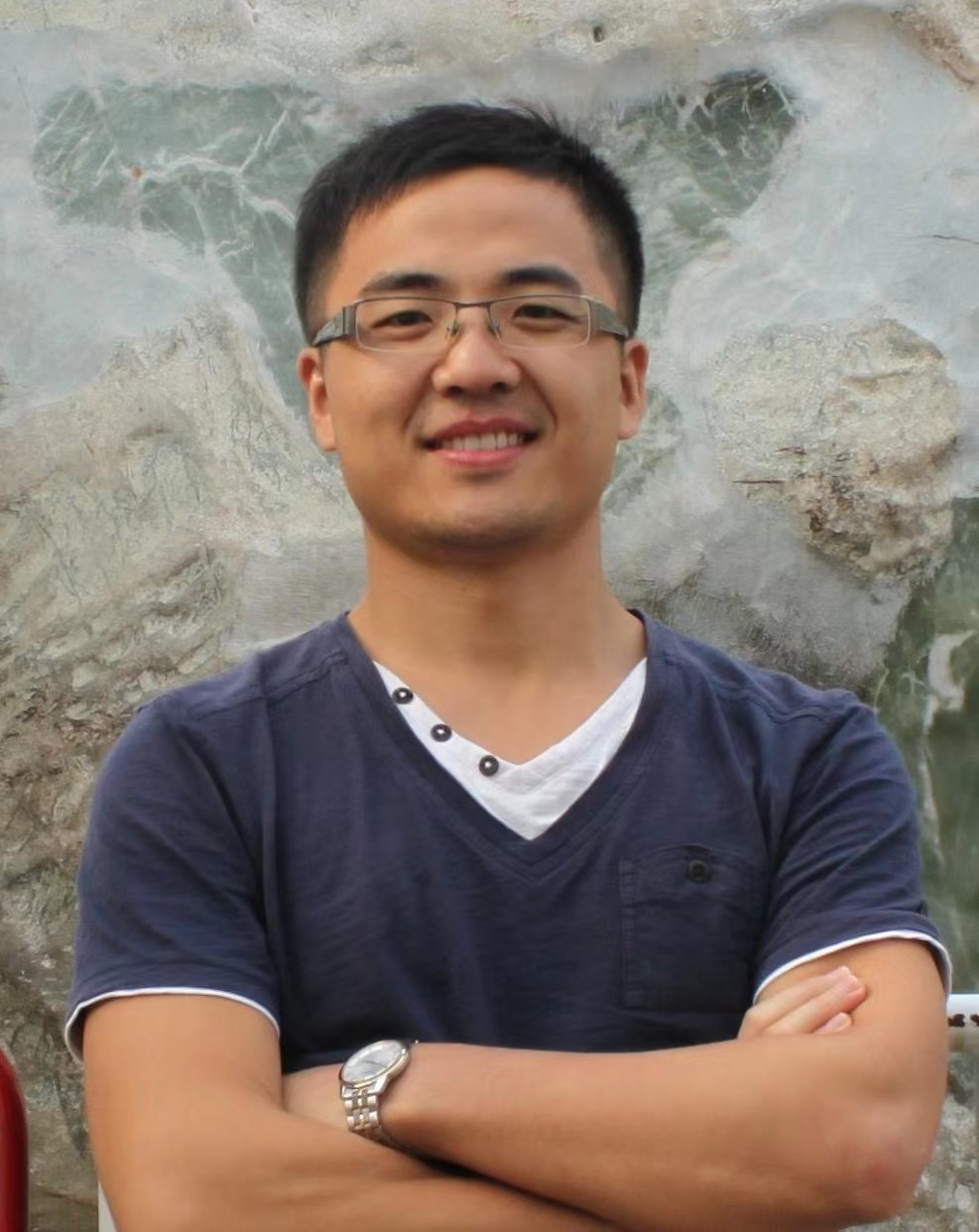}}]{Tao Chen}
(Senior Member, IEEE) received the Ph.D. degree in information engineering from Nanyang Technological University, Singapore, in 2013. He was a Research Scientist with the Institute for Infocomm Research, A*STAR, Singapore, from 2013 to 2017, and a Senior Scientist with the Huawei Singapore Research Center from 2017 to 2018. He is currently a Professor with the College of Future Information Technology, Fudan University, Shanghai, China. His main research interests include efficient computer vision, multimodal visual analysis, large VLM compression, and their applications in embodied robot intelligence, scene understanding, and reconstruction. He has published over 200 papers in international journals and conferences such as IEEE TPAMI, IEEE TIP, IJCV, CVPR, and NeurIPS. He has served in area chair and senior program committee roles for conferences such as AAAI, ICLR, and PRCV. He received the IJCAI 2025 Distinguished Paper Award.
\end{IEEEbiography}

\end{document}

\title{Supplementary Material for: \emph{Advancing MLLM-based UAV Image Understanding and Reasoning: A Benchmark and a Training-Free Multi-Agent System}}

% \author{Haoyu Zhang, Shuoxun Zhang, Peng Ye, Lin Zhang, Jiakang Yuan, Yuening Wang, Feng Xu, and Tao Chen}

\maketitle
\appendices

\section{More details about UAVQA-Bench}
\label{sec:benchmark_appendix}
\subsection{Question Templates}
To ensure the linguistic diversity of UAVQA-Bench, we develop a variety of question templates for each task. This section presents representative templates that define the interaction logic for the agent. 
% \onecolumn
%111
\noindent
\subsubsection{Existence Detection (ED)}
%\begin{itemize}[leftmargin=*, topsep=4pt, itemsep=6pt, parsep=0pt]
\begin{itemize}[leftmargin=*]
    \item \textbf{Scene Presence}
    %\begin{itemize}[label=$\circ$, leftmargin=1.5em, topsep=2pt, itemsep=1pt]
    \begin{itemize}[label=$\circ$, leftmargin=1.5em]
        \item Is any $\{\text{object}\}$ present here?

        \item Is any $\{\text{object}\}$ visible in the image?

        \item Is there any $\{\text{object}\}$ in the scene?
        
        \item Does this image contain any $\{\text{object}\}$?
        
        \item Can you see any $\{\text{object}\}$ in this picture?       
    \end{itemize}
    
    \item \textbf{Conditional Presence} 
    %\begin{itemize}[label=$\circ$, leftmargin=1.5em, topsep=2pt, itemsep=1pt]
    \begin{itemize}[label=$\circ$, leftmargin=1.5em]

        \item Is any $\{\text{object}\}$ present $\{\text{condition}\}$?
        
        \item Is a/an $\{\text{object}\}$ $\{\text{condition}\}$ in the image?
               
        \item Is there any $\{\text{object}\}$ $\{\text{condition}\}$ in this image?

        \item Does the image contain a/an $\{\text{object}\}$ $\{\text{condition}\}$?

        \item Can you see any $\{\text{object}\}$ $\{\text{condition}\}$ in this picture?
        
    \end{itemize}
    
    \item \textbf{Multi-Target Presence} 
    %\begin{itemize}[label=$\circ$, leftmargin=1.5em, topsep=2pt, itemsep=1pt]
    \begin{itemize}[label=$\circ$, leftmargin=1.5em]
        \item Which of the listed objects can be seen in the image?
        \item Does the image show any of the objects listed below?
        \item Which of the following objects are present in the image?
        \item From the options below, which objects appear in the picture?
        \item Among the following choices, which ones are present in the photo?
    \end{itemize}
\end{itemize}

%222
\noindent
\subsubsection{Category Recognition (CR)}
%\begin{itemize}[leftmargin=*, topsep=4pt, itemsep=6pt, parsep=0pt]
\begin{itemize}[leftmargin=*]
    \item \textbf{Regional Classification}
    %\begin{itemize}[label=$\circ$, leftmargin=1.5em, topsep=2pt, itemsep=1pt]
    \begin{itemize}[label=$\circ$, leftmargin=1.5em]

        \item Identify the object found in $\{\mathopen{<}x_1\mathclose{>}\mathopen{<}y_1\mathclose{>}\mathopen{<}x_2\mathclose{>}\mathopen{<}y_2\mathclose{>}\}$.
    
        \item Classify the object in region $\{\mathopen{<}x_1\mathclose{>}\mathopen{<}y_1\mathclose{>}\mathopen{<}x_2\mathclose{>}\mathopen{<}y_2\mathclose{>}\}$.

        \item Which class does the object in $\{\mathopen{<}x_1\mathclose{>}\mathopen{<}y_1\mathclose{>}\mathopen{<}x_2\mathclose{>}\mathopen{<}y_2\mathclose{>}\}$ belong to?
        
        \item What is the category of the object located at $\{\mathopen{<}x_1\mathclose{>}\mathopen{<}y_1\mathclose{>}\mathopen{<}x_2\mathclose{>}\mathopen{<}y_2\mathclose{>}\}$?  

        \item Label the region $\{\mathopen{<}x_1\mathclose{>}\mathopen{<}y_1\mathclose{>}\mathopen{<}x_2\mathclose{>}\mathopen{<}y_2\mathclose{>}\}$ with the correct object category.
    \end{itemize}
\end{itemize}

%333
\noindent
\subsubsection{Quantity Awareness (QA)}
%\begin{itemize}[leftmargin=*, topsep=4pt, itemsep=6pt, parsep=0pt]
\begin{itemize}[leftmargin=*]
    \item \textbf{Scene Counting}
    %\begin{itemize}[label=$\circ$, leftmargin=1.5em, topsep=2pt, itemsep=1pt]
    \begin{itemize}[label=$\circ$, leftmargin=1.5em]
        \item Provide the quantity of $\{\text{object}\}$. 
        
        \item Count all the $\{\text{object}\}$ in this picture.
              
        \item How many $\{\text{object}\}$ are in this image?
        
        \item How many instances of $\{\text{object}\}$ can you see?

        \item Can you count the number of $\{\text{object}\}$ present?
          
    \end{itemize}
    
    \item \textbf{Regional Counting} 
    %\begin{itemize}[label=$\circ$, leftmargin=1.5em, topsep=2pt, itemsep=1pt]
    \begin{itemize}[label=$\circ$, leftmargin=1.5em]

        \item Provide the quantity of $\{\text{object}\}$ $\{\text{condition}\}$.
           
        \item Count all the $\{\text{object}\}$ $\{\text{condition}\}$ in this picture.
               
        \item What is the total number of $\{\text{object}\}$ visible $\{\text{condition}\}$?

        \item How many $\{\text{object}\}$ appear inside this bounding box $\{\mathopen{<}x_1\mathclose{>}\mathopen{<}y_1\mathclose{>}\mathopen{<}x_2\mathclose{>}\mathopen{<}y_2\mathclose{>}\}$?

        \item What is the total number of $\{\text{object}\}$ in the bounding box $\{\mathopen{<}x_1\mathclose{>}\mathopen{<}y_1\mathclose{>}\mathopen{<}x_2\mathclose{>}\mathopen{<}y_2\mathclose{>}\}$?
                
        \item Can you count the number of $\{\text{object}\}$ within the bounding box $\{\mathopen{<}x_1\mathclose{>}\mathopen{<}y_1\mathclose{>}\mathopen{<}x_2\mathclose{>}\mathopen{<}y_2\mathclose{>}\}$?
        
    \end{itemize}
    
    \item \textbf{Number Comparison} 
    %\begin{itemize}[label=$\circ$, leftmargin=1.5em, topsep=2pt, itemsep=1pt]
    \begin{itemize}[label=$\circ$, leftmargin=1.5em]

        \item Do $\{\text{object}\}_A$ outnumber $\{\text{object}\}_B$?
        
        \item Do $\{\text{object}\}_A$ exceed $\{\text{object}\}_B$ in quantity?

        \item Are $\{\text{object}\}_A$ less abundant than $\{\text{object}\}_B$?
        
        \item Are $\{\text{object}\}_A$ more numerous than $\{\text{object}\}_B$?
        
        \item Is it true that $\{\text{object}\}_A$ is fewer than $\{\text{object}\}_B$?
        
    \end{itemize}
\end{itemize}

%444
\noindent
\subsubsection{Fine-grained Attribute Perception (FAP)}
%\begin{itemize}[leftmargin=*, topsep=4pt, itemsep=6pt, parsep=0pt]
\begin{itemize}[leftmargin=*]
    \item \textbf{Regional Attribute Recognition}
    %\begin{itemize}[label=$\circ$, leftmargin=1.5em, topsep=2pt, itemsep=1pt]
    \begin{itemize}[label=$\circ$, leftmargin=1.5em]
    
        \item What color is the $\{\text{object}\}$ in $\{\mathopen{<}x_1\mathclose{>}\mathopen{<}y_1\mathclose{>}\mathopen{<}x_2\mathclose{>}\mathopen{<}y_2\mathclose{>}\}$?

        \item Identify the color of the $\{\text{object}\}$ in $\{\mathopen{<}x_1\mathclose{>}\mathopen{<}y_1\mathclose{>}\mathopen{<}x_2\mathclose{>}\mathopen{<}y_2\mathclose{>}\}$.
        
        \item Which color does the $\{\text{object}\}$ in $\{\mathopen{<}x_1\mathclose{>}\mathopen{<}y_1\mathclose{>}\mathopen{<}x_2\mathclose{>}\mathopen{<}y_2\mathclose{>}\}$ have?

        \item Identify the driving direction (relative to the camera) of the $\{\text{object}\}$ in $\{\mathopen{<}x_1\mathclose{>}\mathopen{<}y_1\mathclose{>}\mathopen{<}x_2\mathclose{>}\mathopen{<}y_2\mathclose{>}\}$.
        
        \item What is the driving direction (relative to the camera) of the $\{\text{object}\}$ located at $\{\mathopen{<}x_1\mathclose{>}\mathopen{<}y_1\mathclose{>}\mathopen{<}x_2\mathclose{>}\mathopen{<}y_2\mathclose{>}\}$?

    \end{itemize}
    
    \item \textbf{Function Recognition} 
    %\begin{itemize}[label=$\circ$, leftmargin=1.5em, topsep=2pt, itemsep=1pt]
    \begin{itemize}[label=$\circ$, leftmargin=1.5em]

        \item What is the key function of the $\{\text{object}\}$ bounded by $\{\mathopen{<}x_1\mathclose{>}\mathopen{<}y_1\mathclose{>}\mathopen{<}x_2\mathclose{>}\mathopen{<}y_2\mathclose{>}\}$?
        \item Which functional category does the $\{\text{object}\}$ in $\{\mathopen{<}x_1\mathclose{>}\mathopen{<}y_1\mathclose{>}\mathopen{<}x_2\mathclose{>}\mathopen{<}y_2\mathclose{>}\}$ belong to?
        \item The $\{\text{object}\}$ within the bounding box $\{\mathopen{<}x_1\mathclose{>}\mathopen{<}y_1\mathclose{>}\mathopen{<}x_2\mathclose{>}\mathopen{<}y_2\mathclose{>}\}$ falls under which functional category?

    \end{itemize}
    
    \item \textbf{Safety Landing} 
    %\begin{itemize}[label=$\circ$, leftmargin=1.5em, topsep=2pt, itemsep=1pt]
    \begin{itemize}[label=$\circ$, leftmargin=1.5em]

        \item Identify the safest landing area for a drone among the given options.
        \item Among the options below, where is the best place for a drone to land?
        \item Select the optimal landing spot for a drone from the following choices.
        \item Which location is most ideal for a drone landing in 
        the options provided?
        \item From the following, choose the most appropriate site for a drone to land.
        
    \end{itemize}
\end{itemize}

%555
\noindent
\subsubsection{Spatial Relationship Understanding (SRU)}
%\begin{itemize}[leftmargin=*, topsep=4pt, itemsep=6pt, parsep=0pt]
\begin{sloppypar}
\begin{itemize}[leftmargin=*]
    \item \textbf{Spatial Relation}
    %\begin{itemize}[label=$\circ$, leftmargin=1.5em, topsep=2pt, itemsep=1pt]
    \begin{itemize}[label=$\circ$, leftmargin=1.5em]

        \item Relative to the object in $\{\mathopen{<}x_1\mathclose{>}\mathopen{<}y_1\mathclose{>}\mathopen{<}x_2\mathclose{>}\mathopen{<}y_2\mathclose{>}\}_A$, where is the object in $\{\mathopen{<}x_1\mathclose{>}\mathopen{<}y_1\mathclose{>}\mathopen{<}x_2\mathclose{>}\mathopen{<}y_2\mathclose{>}\}_B$?

        \item As seen by the object in $\{\mathopen{<}x_1\mathclose{>}\mathopen{<}y_1\mathclose{>}\mathopen{<}x_2\mathclose{>}\mathopen{<}y_2\mathclose{>}\}_A$, in what direction is the object in $\{\mathopen{<}x_1\mathclose{>}\mathopen{<}y_1\mathclose{>}\mathopen{<}x_2\mathclose{>}\mathopen{<}y_2\mathclose{>}\}_B$?

        \item From the viewpoint of the object in $\{\mathopen{<}x_1\mathclose{>}\mathopen{<}y_1\mathclose{>}\mathopen{<}x_2\mathclose{>}\mathopen{<}y_2\mathclose{>}\}_A$, what is the direction to the object in $\{\mathopen{<}x_1\mathclose{>}\mathopen{<}y_1\mathclose{>}\mathopen{<}x_2\mathclose{>}\mathopen{<}y_2\mathclose{>}\}_B$?
              
        \item Assuming the UAV's forward direction corresponds to the top of the image, at which clock position is the object in $\{\mathopen{<}x_1\mathclose{>}\mathopen{<}y_1\mathclose{>}\mathopen{<}x_2\mathclose{>}\mathopen{<}y_2\mathclose{>}\}$ located?
        
        \item Given that the top of the image is treated as the front of the UAV, what is the clock position of the object within $\{\mathopen{<}x_1\mathclose{>}\mathopen{<}y_1\mathclose{>}\mathopen{<}x_2\mathclose{>}\mathopen{<}y_2\mathclose{>}\}$ relative to the UAV?         

    \end{itemize}
    
    \item \textbf{Height Comparison} 
    %\begin{itemize}[label=$\circ$, leftmargin=1.5em, topsep=2pt, itemsep=1pt]
    \begin{itemize}[label=$\circ$, leftmargin=1.5em]
       \item Which between $\{\mathopen{<}x_1\mathclose{>}\mathopen{<}y_1\mathclose{>}\mathopen{<}x_2\mathclose{>}\mathopen{<}y_2\mathclose{>}\}_A$ and $\{\mathopen{<}x_1\mathclose{>}\mathopen{<}y_1\mathclose{>}\mathopen{<}x_2\mathclose{>}\mathopen{<}y_2\mathclose{>}\}_B$ is determined to be at a higher position?

        \item Between the two positions $\{\mathopen{<}x_1\mathclose{>}\mathopen{<}y_1\mathclose{>}\mathopen{<}x_2\mathclose{>}\mathopen{<}y_2\mathclose{>}\}_A$ and $\{\mathopen{<}x_1\mathclose{>}\mathopen{<}y_1\mathclose{>}\mathopen{<}x_2\mathclose{>}\mathopen{<}y_2\mathclose{>}\}_B$, which one is judged to be higher in the frame?
        
        \item Please judge which object is higher in the frame between $\{\mathopen{<}x_1\mathclose{>}\mathopen{<}y_1\mathclose{>}\mathopen{<}x_2\mathclose{>}\mathopen{<}y_2\mathclose{>}\}_A$ and $\{\mathopen{<}x_1\mathclose{>}\mathopen{<}y_1\mathclose{>}\mathopen{<}x_2\mathclose{>}\mathopen{<}y_2\mathclose{>}\}_B$?
        \item Can you judge which contains an object at a higher position between $\{\mathopen{<}x_1\mathclose{>}\mathopen{<}y_1\mathclose{>}\mathopen{<}x_2\mathclose{>}\mathopen{<}y_2\mathclose{>}\}_A$ and $\{\mathopen{<}x_1\mathclose{>}\mathopen{<}y_1\mathclose{>}\mathopen{<}x_2\mathclose{>}\mathopen{<}y_2\mathclose{>}\}_B$?

        \item Please evaluate the two positions $\{\mathopen{<}x_1\mathclose{>}\mathopen{<}y_1\mathclose{>}\mathopen{<}x_2\mathclose{>}\mathopen{<}y_2\mathclose{>}\}_A$ and $\{\mathopen{<}x_1\mathclose{>}\mathopen{<}y_1\mathclose{>}\mathopen{<}x_2\mathclose{>}\mathopen{<}y_2\mathclose{>}\}_B$, and indicate which one is higher in the vertical direction.
        
    \end{itemize}
    
    \item \textbf{Distance Comparison} 
    %\begin{itemize}[label=$\circ$, leftmargin=1.5em, topsep=2pt, itemsep=1pt]
    \begin{itemize}[label=$\circ$, leftmargin=1.5em]
  
        \item Which between $\{\mathopen{<}x_1\mathclose{>}\mathopen{<}y_1\mathclose{>}\mathopen{<}x_2\mathclose{>}\mathopen{<}y_2\mathclose{>}\}_A$ and $\{\mathopen{<}x_1\mathclose{>}\mathopen{<}y_1\mathclose{>}\mathopen{<}x_2\mathclose{>}\mathopen{<}y_2\mathclose{>}\}_B$ is measured to be closer to the shooting point?
        
        \item Between $\{\mathopen{<}x_1\mathclose{>}\mathopen{<}y_1\mathclose{>}\mathopen{<}x_2\mathclose{>}\mathopen{<}y_2\mathclose{>}\}_A$ and $\{\mathopen{<}x_1\mathclose{>}\mathopen{<}y_1\mathclose{>}\mathopen{<}x_2\mathclose{>}\mathopen{<}y_2\mathclose{>}\}_B$, which object has a smaller depth value, meaning it is closer?

        \item Please evaluate the two positions $\{\mathopen{<}x_1\mathclose{>}\mathopen{<}y_1\mathclose{>}\mathopen{<}x_2\mathclose{>}\mathopen{<}y_2\mathclose{>}\}_A$ and $\{\mathopen{<}x_1\mathclose{>}\mathopen{<}y_1\mathclose{>}\mathopen{<}x_2\mathclose{>}\mathopen{<}y_2\mathclose{>}\}_B$, and indicate which one is closer to the drone.
        
        \item Please judge which object is closer to the drone between $\{\mathopen{<}x_1\mathclose{>}\mathopen{<}y_1\mathclose{>}\mathopen{<}x_2\mathclose{>}\mathopen{<}y_2\mathclose{>}\}_A$ and $\{\mathopen{<}x_1\mathclose{>}\mathopen{<}y_1\mathclose{>}\mathopen{<}x_2\mathclose{>}\mathopen{<}y_2\mathclose{>}\}_B$?

    \end{itemize}
\end{itemize}
\end{sloppypar}

%666
\noindent
\subsubsection{Visual Grounding (VG)}
%\begin{itemize}[leftmargin=*, topsep=4pt, itemsep=6pt, parsep=0pt]
\begin{itemize}[leftmargin=*]
    \item \textbf{Simple Object Grounding}
    %\begin{itemize}[label=$\circ$, leftmargin=1.5em, topsep=2pt, itemsep=1pt]
    \begin{itemize}[label=$\circ$, leftmargin=1.5em]

        \item Locate the $\{\text{object}\}$.

        \item Point out the $\{\text{object}\}$.

        \item Could you point out the $\{\text{object}\}$?
            
        \item Draw the position of a/an $\{\text{object}\}$.

        \item Can you show me where the $\{\text{object}\}$ is?
        
    \end{itemize}
    
    \item \textbf{Complex Semantic Grounding} 
    %\begin{itemize}[label=$\circ$, leftmargin=1.5em, topsep=2pt, itemsep=1pt]
    \begin{itemize}[label=$\circ$, leftmargin=1.5em]

       \item Please give me the location of $\{\text{semantics}\}$.
       \item Please indicate the position of $\{\text{semantics}\}$.
       \item Can you point out the location of $\{\text{semantics}\}$?
       \item Could you tell me the location for $\{\text{semantics}\}$?     
       \item In the current image, where can I find $\{\text{semantics}\}$?
        
    \end{itemize}
    
    \item \textbf{Highest Object Grounding} 
    %\begin{itemize}[label=$\circ$, leftmargin=1.5em, topsep=2pt, itemsep=1pt]
    \begin{itemize}[label=$\circ$, leftmargin=1.5em]
  
        \item Find the area with the tallest building.

        \item Which area in the image has the tallest building?
        \item Judge which building area in the image has the highest floors.

    \end{itemize}
\end{itemize}

\subsubsection{Template Visualizations}
\begin{figure}[!ht]
\centering
\includegraphics[width=0.7\linewidth]{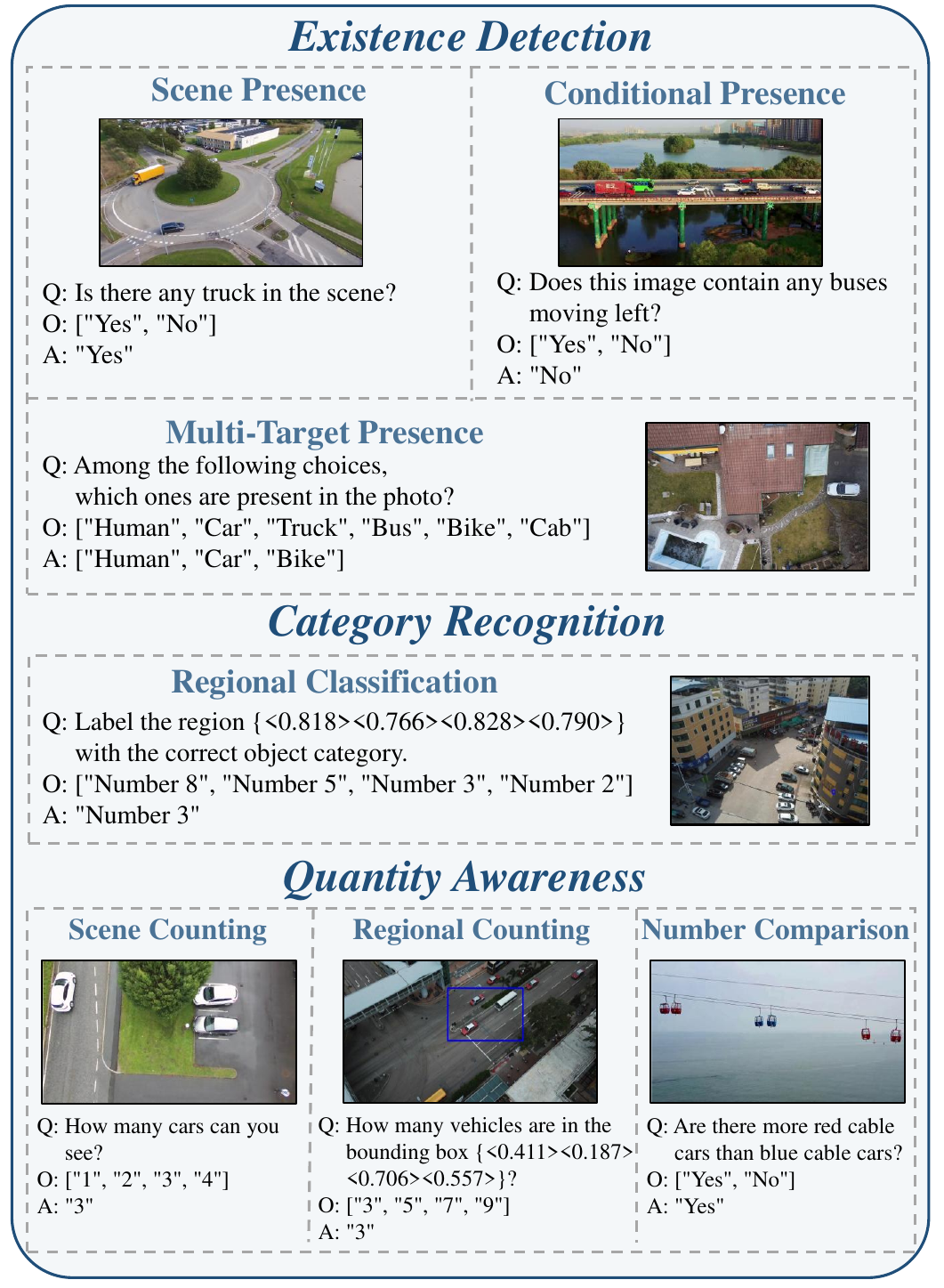}
\caption{Sample instances for tasks in Existence Detection, Category Recognition, and Quantity Awareness.} 
\label{task1}
\end{figure}

\begin{figure}[!t]
\centering
\includegraphics[width=0.7\linewidth]{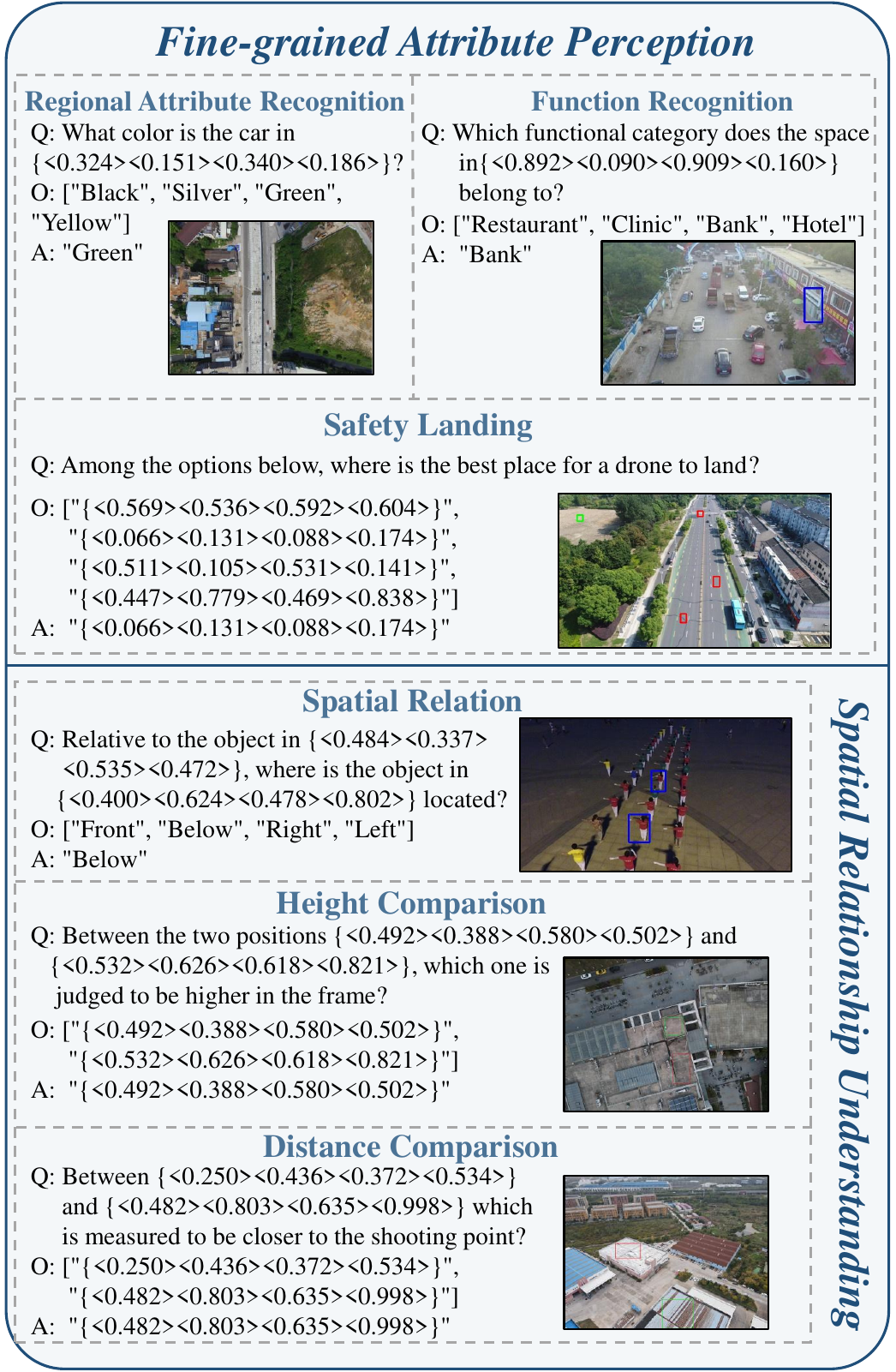}
\caption{Sample instances for tasks in Fine-grained Attribute Perception and Spatial Relationship Understanding.} 
\label{task2}
\end{figure}

\begin{figure}[!t]
\centering
\includegraphics[width=0.7\linewidth]{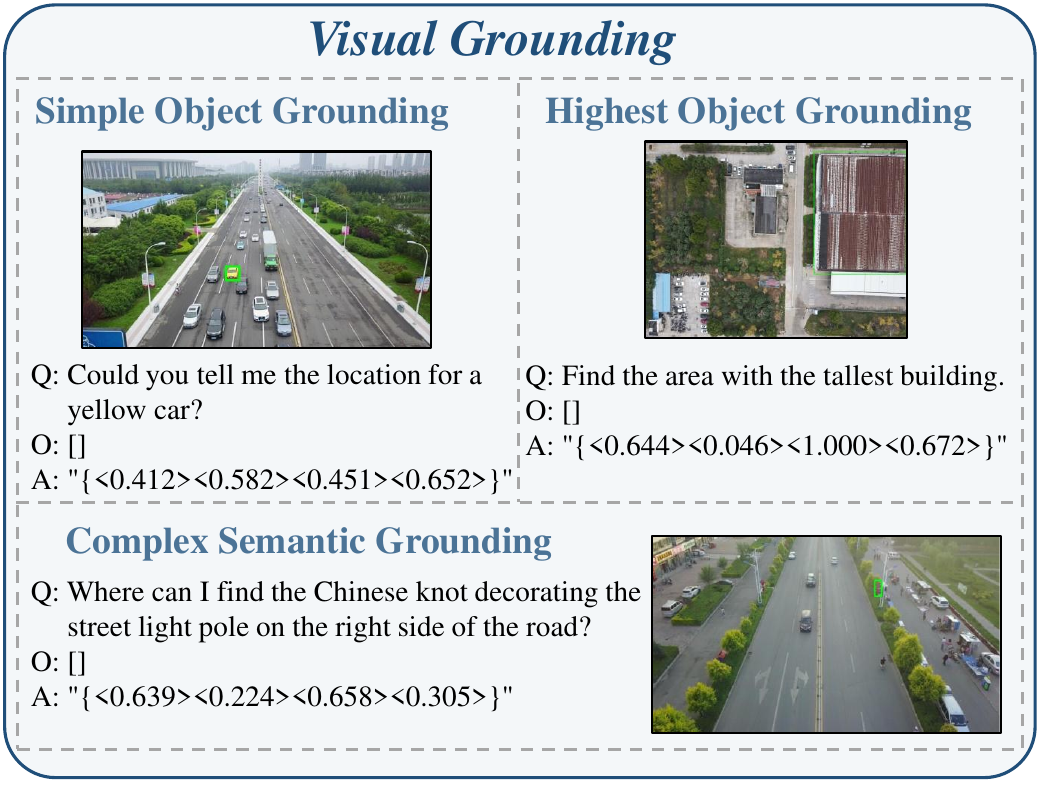}
\caption{Sample instances for tasks in Visual Grounding.} 
\label{task3}
\end{figure}

In this section, we present additional examples from UAVQA-Bench. 
% Sample instances and visual prompts for each task are illustrated in Figures , \ref{task2}, and \ref{task3}.
% Figure~\ref{task1} provides sample instances and visual prompts for the tasks of Existence Detection, Category Recognition, and Quantity Awareness. Figure~\ref{task2} illustrates examples for Fine-grained Attribute Perception and Spatial Relationship Understanding. Finally, Figure ~\ref{task3} showcases samples for the Visual Grounding task. In these samples, ``Q'', ``O'', and ``A'' denote the Question, Options, and Ground-truth Answer, respectively.
Figure~\ref{task1} provides sample instances and visual prompts for the tasks of Existence Detection, Category Recognition, and Quantity Awareness. Figure~\ref{task2} illustrates examples for Fine-grained Attribute Perception and Spatial Relationship Understanding. Finally, Figure~\ref{task3} showcases samples for the Visual Grounding tasks. In these samples, ``Q'', ``O'', and ``A'' denote the Question, Options, and Ground-truth Answer, respectively. For clarity, the sample figures include bounding boxes in different colors: red boxes indicate incorrect candidate answers, green boxes denote the correct answer, and blue boxes mark the object or region referred to in the question. These bounding boxes are shown only for visualization in the paper and are not present in the actual images provided to the model during inference.

\clearpage
% \twocolumn
\section{More Details of Our Method}
\label{sec:appendix_method_details}
In this section, we provide the implementation settings and agent configurations used in our experiments.

\subsection{Implementation Details}
All MLLM calls use a temperature of $0.7$ and a maximum of 8,192 new tokens. For DAAS, the maximum depth is $D=5$, and the search width is set adaptively to $W=\min(3,|\mathcal{T}_{\mathrm{opt}}|)$, where $|\mathcal{T}_{\mathrm{opt}}|$ is the number of tools selected by DSPE. Direct MLLM baselines use their default system prompts with the same image--query input structure, temperature, and token budget. All experiments are conducted on NVIDIA H200 GPUs.

\subsection{Agent Configurations}
We detail the configuration of each agent in our framework in Table~\ref{tab:agent_configuration}. We utilize the Qwen3-VL series as the backbone MLLM. Specifically, we employ the \textit{Instruct} variant for agents that require precise execution of specialized instructions (e.g., $\text{Agent}_{\text{TS}}$, $\text{Agent}_{\text{SR}}$), and the \textit{Thinking} variant for the Long-Chain Answer Agent ($\text{Agent}_{\text{LA}}$) to leverage its enhanced capabilities in complex reasoning and synthesis. It is worth noting that the \textit{Instruct} and \textit{Thinking} models share the identical architectural design, differing only in their parameter weights which are tuned for instruction compliance and deep reasoning, respectively.

\begin{table}[ht]
    \centering
    \caption{Agent configurations and model types used in UAV-MAS.}
    \label{tab:agent_configuration}
    \begin{tabular}{l|c}
    \toprule
    \textbf{Agent Role} & \textbf{Model Type}\\
    \midrule
    Tool Select Agent ($\text{Agent}_{\text{TS}}$)& Qwen3-VL Instruct \\
    Strategic Reasoning Agent ($\text{Agent}_{\text{SR}}$) & Qwen3-VL Instruct\\
    Perceptual Verification Agent ($\text{Agent}_{\text{PV}}$) & Qwen3-VL Instruct\\
    Contextual Integration Agent ($\text{Agent}_{\text{CI}}$) & Qwen3-VL Instruct \\
    Long-Chain Answer Agent ($\text{Agent}_{\text{LA}}$) & Qwen3-VL Thinking\\
    Score Agent ($\text{Agent}_{\text{SC}}$) & Qwen3-VL Instruct\\
    \bottomrule
    \end{tabular}
\end{table}